\documentclass{article}

\usepackage[final]{corl_2018} % Uncomment for the camera-ready ``final'' version

\usepackage{amsmath}
\usepackage{amsfonts}
\DeclareMathOperator{\E}{\mathbb{E}}
\usepackage{xcolor}

\usepackage{graphicx}
\usepackage{subfigure}
\usepackage{caption}
\usepackage[ruled]{algorithm2e}
\usepackage{float}
\usepackage{comment}
\usepackage{multicol}
\usepackage{enumitem}
\usepackage{multirow}

\author{
  Zih-Yun Chiu\\
  National Taiwan University\\
  \texttt{z.y.sarah.chiu@gmail.com} \\
  \And
  Yi-Lin Tuan\\
  National Taiwan University\\
  \texttt{pascaltuan@gmail.com} \\
  \AND
  Hung-yi Lee\\
  National Taiwan University\\
  \texttt{hungyilee@ntu.edu.tw} \\
  \And
  Li-Chen Fu\\
  National Taiwan University\\
  \texttt{lichen@ntu.edu.tw} \\
}

\begin{document}
\title{Parallelized Reverse Curriculum Generation}

\maketitle

%===============================================================================

\begin{abstract}
For reinforcement learning (RL), it is challenging for an agent to master a task that requires a specific series of actions due to sparse rewards.
To solve this problem, reverse curriculum generation (RCG) provides a reverse expansion approach that automatically generates a curriculum for the agent to learn.
More specifically, RCG adapts the initial state distribution from the neighborhood of a goal to a distance as training proceeds.
However, the initial state distribution generated for each iteration might be biased, thus making the policy overfit or slowing down the reverse expansion rate.
While training RCG for actor-critic (AC) based RL algorithms, this poor generalization and slow convergence might be induced by the tight coupling between an AC pair.
Therefore, we propose a parallelized approach that simultaneously trains multiple AC pairs and periodically exchanges their critics.
We empirically demonstrate that this proposed approach can improve RCG in performance and convergence, and it can also be applied to other AC based RL algorithms with adapted initial state distribution.
\end{abstract}

%===============================================================================

\section{Introduction}
\label{intro}
In robotics, deep RL has recently achieved promising results for many applications, including manipulation~\citep{popov2017data}, locomotion~\citep{peng2017deeploco}, and navigation~\citep{zhu2017target}.
For these applications, how the reward functions are designed will dramatically influence the performance of the agents since the learning of RL agents is driven by the reward functions.
A non-sparse reward function can make the optimization process easier, but the learned policy might generate actions that are short-sighted~\citep{florensa2017reverse}.
This is problematic when an agent has to {\it reach the final goal by performing a series of actions} (e.g., Ring on Peg task~\citep{florensa2017reverse}, clean-up task~\citep{riedmiller2018learning}). For these tasks, a sparse reward function seems to be a better choice. 

However, using a sparse reward function will make the task difficult to learn since the agent can seldom reach the goal via a simple exploration process.
To deal with this difficulty, several methods have been proposed, including learning from demonstration~\citep{smart2002effective,vevcerik2017leveraging}, simulation-to-reality transfer learning~\citep{rusu2016sim}, and curriculum-based approaches~\citep{andreas2016modular,narvekar2016source}.
Most of the methods mentioned above require prior knowledge provided by humans.
Recent breakthroughs have been made with limited prior knowledge, e.g., hindsight experience replay (HER)~\citep{andrychowicz2017hindsight}, scheduled auxiliary control (SAC-X) \citep{riedmiller2018learning}, and reverse curriculum generation for RL~\citep{florensa2017reverse}.

Among these methods, reverse curriculum generation (RCG) requires only the prior knowledge of states near success, and utilizes dynamic programming to start learning the task from states near the goal, gradually expanding to distant states.
In detail, the states that are selected from a given initial state distribution and with some chance to reach the goal are then utilized to expand the initial state distribution through the \emph{sampling nearby process}. 
There exist some potential issues for this algorithm:  
The initial state distribution might only cover part of the feasible state space.
This makes the policy overfit, or the initial state distribution might take a long time to expand to the target one.

In this paper, we propose parallelized reverse curriculum generation (PRCG) which can improve the performance and convergence of RCG.
%When using RCG for an AC framework in RL, we observe tight coupling, that is, \red{the learned policy and value highly depend on each other.}
When using RCG for an AC framework in RL, we observe that the initial states sampled will largely depend on the value predicted by the critic.
As a result, the actor might only learn from a biased initial state distribution.
This leads to overfitting and slows down the exploration of both the actor and the critic .
To reduce this issue, PRCG simultaneously trains multiple AC models and randomly exchanges their critics every $K$ iterations.
The experiments show that since each agent can learn from more than one critic, there is a higher probability for each model to learn a more general policy and thus improve the performance and exploration ability.

Training multiple agents simultaneously to improve learning is not a new idea, but we observe that our method is easier to find a set of good hyperparameters than an asynchronous method like A3C~\citep{mnih2016asynchronous}.
Moreover, although we primarily apply our proposed parallelized approach to RCG, its effect is not limited to this algorithm.
In fact, since this approach not only prevents overfitting but enlarges environmental information by exchanging critics, it can benefit any AC based RL algorithm with adapted initial state distribution.
In our experiments, we demonstrate: (1) PRCG outperforms RCG as well as an ensemble and an asynchronous versions of RCG; 
(2) although there are other strategies that can prevent overfitting and increase exploration, (e.g., swapping or sharing initial states), swapping critics achieves better exploration and stability;
(3) our proposed parallelized approach can improve the performance and accelerate the training on a widely used AC based RL algorithm with adapted initial state distribution.

In conclusion, the main contribution of this paper includes:

\begin{enumerate}[leftmargin=*]
    \item We indicate the potential problems of RCG, and demonstrate that these problems lead to overfitting and limited exploration.
    \item We propose a parallelized training approach to facilitate RCG. This method can increase the generalization ability and exploration of the model.
    \item We demonstrate that this parallelized approach can also improve other AC based RL algorithms with adapted initial state distribution.
\end{enumerate}

\section{Motivation}
\label{sec:motivation}

The parallelized approach we propose for PRCG simultaneously trains multiple AC models and randomly exchanges their critics during training. 
This idea is actually inspired by a method of solving model collapse for generative adversarial networks (GANs)~\citep{goodfellow2016nips}. 
In this section, we will briefly explain the similarities between the issues in RCG and mode collapse in GANs as well as how PRCG can benefit from these similarities.

In GANs, a generator aims to generate authentic samples that mimic the real data distribution while a discriminator aims to judge if a sample is from the real data distribution or generated by the generator.
The generator and the discriminator are adversaries, which means that they optimize opposite objectives.
\emph{Mode collapse} is one of the well-known issues in GANs. 
A real data distribution in GANs can possibly be divided into several parts, and here we call each part a \emph{mode}.
If a generator is trained to fit a mode well, it might stop fitting other modes since it can already fool the discriminator.
Hence, this generator will generate similar samples and fail to describe the whole real data distribution.
This problem is called \emph{mode collapse} in GANs.

In AC methods, given a state, an actor predicts an action that leads the agent to another state, and a critic evaluates the state. 
The actor and critic are then updated based on the rewards from the environment and the outputs of the critic.
Although GANs and AC methods both contain two dependent models, unlike in GANs, an actor and a critic in AC methods collaborate with each other to achieve a good policy. 
Therefore, GANs and general AC methods have a fundamental difference. 
However, when it comes to RCG for AC based RL, this AC method has some similarities with GANs.

This similarity lies in the mechanism of these two models. 
Here we briefly mention the mechanism and issue of RCG for AC methods and will provide a thorough explanation in Section \ref{sec:method}. 
In RCG for AC methods, the critic serves as a filter of the initial states. 
That is, the initial states are not always uniformly sampled, and their distribution largely depends on the critic. 
This chosen initial state distribution will then be slightly expanded by the sampling nearby process. 
Hence, if the critic is not well learned, the initial state distribution in each training iteration might be in an undesired region and very biased. 
This biased initial state distribution will make the actor \emph{overfit} and lead to poor performance. 
The whole initial state space can be divided into several parts, and we also call each part a \emph{mode}.

\begin{table}[t]
    \centering
    \begin{tabular}{r|l|l}
        & GANs & RCG for AC methods \\\hline\hline
        %\multirow{2}{*}{Discriminator/Critic} & estimates the boundaries & selects \emph{good} initial states\\
        %& of real/fake distribution &\\\hline
        \multirow{2}{*}{Discriminator/Critic} & estimates the boundaries & estimates the boundaries\\
        & of real/fake distribution & of initial state distribution\\\hline
        %\multirow{3}{*}{Generator/Actor} & learns the real distribution & learns policy based on \\
        %& based on the estimated & the selected initial states\\
        %& boundaries of real/fake dist. &\\\hline
        \multirow{4}{*}{Generator/Actor} & learns to adapt to & learns to adapt to \\
        & the real data distribution & the initial state distribution \\
        & whose boundaries are estimated & whose boundaries are \\
        & by the discriminator & estimated by the critic\\\hline
        \multirow{2}{*}{\emph{Mode} (our definition)} & the partitions of & the partitions of the desired \\
        & the real data distribution & initial state distribution\\
        \hline
        %\emph{Mode Collapse}/\emph{Overfit} & generator adapts to few modes & actor adapts to few modes\\
        Mode Collapse & generator adapts to & actor adapts to \\
        /Overfitting & a small number of modes & a small number of modes\\
    \end{tabular}
    \vspace{10pt}
    \caption{The similarities between GANs and RCG for AC methods}
    \vspace{-20pt}
    \label{table:gan_ac_similarity}
\end{table}

Based on the mechanism of GANs and RCG for AC methods, Table \ref{table:gan_ac_similarity} shows the similarities between these two models.  
Both a discriminator and a critic estimate the boundaries of the target distribution. 
The target distribution in GANs is the real data distribution; the target distribution in RCG is the initial state distribution. 
Both a generator and an actor aim to adapt to the target distribution which boundaries are estimated by the discriminator/critic. 
Both mode collapse and overfitting happen because the generator/actor only learn a small number of modes. 
These similarities motivate us to utilize an approach of solving mode collapse in RCG for AC methods, and we later generalize this approach to other AC methods with adapted initial state distribution.

%Since GAN and AC methods both contain two dependent models, previous work has connected these two algorithms~\citep{pfau2016connecting}.
%In GANs, a generator produces a sample for a discriminator to judge. Similarly, in an AC method, the actor predicts an action. This action leads the agent to a state, and this state is evaluated by the critic. Their two dependent models are then iteratively trained.
%\red{
%The generator in a GAN aims to generate authentic samples that mimic the real data distribution.
%The actor in an AC method aims to adopt policy that performs the best on the give states.
%}

%\red{
%Particularly, RCG for RL has a similar problem to the \emph{mode collapse} in GAN.
%In GAN, a real data distribution contains several separated modes.
%When a generator reach a mode, it tends to stuck in this local minimum because it can fool the discriminator by sampling from this mode.
%The generator then fails to learn other modes to better describe the whole real data distribution.
%This problem is called \emph{mode collapse} in GAN.
%In RCG for RL, we can interpret the whole possible initial states as a composition of multiple modes.
%When a actor-critic pair learns well in a mode, it also tends to stuck in this local minimum because the critic can only %filter initial states distribution that has been proved good.
%The actor then can only learn policy based on the learned initial states distribution, and fail to explore more.
%}

\section{Related Work}
\label{sec:relate}

\subsection{Mode Collapse in GANs}
In our work, we utilize an approach of solving mode collapse in GANs to solve overtting in RCG for AC methods.
Recently, there are several methods proposed to solve mode collapse, including batch normalization~\citep{radford2015unsupervised}, minibatch normalization~\citep{salimans2016improved}, Unrolled GANs~\citep{metz2016unrolled}, Stack GANs~\citep{huang2017stacked}, VEEGAN~\citep{srivastava2017veegan}, GAP~\citep{im2016generative}, D2GAN~\citep{nguyen2017dual}, and E-GAN~\citep{wang2018evolutionary}.
GAP, D2GAN and E-GAN all use more than one generator or discriminator to achieve stable training and improve performance. 
%Although mode collapse is still an open issue for GANs, these methods provide an insight of its origin or a solution to it.
These methods all provide an insight or a solution to mode collapse.
%Since the problem that GAP aims to solve is similar to ours, we borrow the idea of GAP, which is to train multiple AC pairs simultaneously and exchange their critics during the training process.
Since the idea behind GAP is more similar to what we want to achieve, we borrow the idea of GAP, which is to train multiple GANs simultaneously and exchange their discriminators during the training process.

\subsection{Parallelized Methods in RL}
Previous research has worked on simultaneously training multiple agents.
Distributed deep Q-learning~\citep{ong2015distributed} and distributed proximal policy optimization (DPPO)~\citep{heess2017emergence} distribute data collection and gradients calculation over workers.
The general reinforcement learning architecture (Gorilla)~\citep{nair2015massively} and asynchronous advantage actor-critic (A3C)~\citep{mnih2016asynchronous} let multiple agents interact with their own environments and send the gradients to the central learner. The learned parameters are sent back to each agent at regular intervals.
Most of the parallelized RL methods mentioned above need a central network to collect the gradients from each local network, and the parameters of local networks are periodically set to the same. 
In contrast, in our method, each agent only shares their information by exchanging critics, and the network parameters of each model can become diverse during the learning process.
We will demonstrate that while other parallelized RL methods periodically synchronize all the models to the same, each model in our proposed approach develops its own strategy, thus solving the problems of RCG.

\section{Background}
\label{sec:back}

In this section, we will formulate a RL problem, and introduce RCG~\citep{florensa2017reverse} and GAP~\citep{im2016generative}.

\subsection{Preliminaries}
\label{subsec:pre}
For the RL problem of an MDP, we utilize the following definitions: Let $M = (S, A, P, r, \rho_0, \gamma, T)$ be the tuple for the discrete-time finite-horizon MDP, where $S$ is the state set, $A$ is the action set, $P$ is the transition probability distribution, $r$ is the reward function, $\rho_0$ is the initial state distribution, $\gamma$ is the discount factor, and $T$ is the horizon.
The stochastic policy $\pi_\theta$ parameterized by $\theta$ is learned to maximize the expected return $\E_{\pi_{\theta}(a|s), s_0\sim\rho_0}[R_{s_0}(\tau)]$. The sum of discounted rewards for a whole trajectory $\tau = (s_0, a_0, \cdots, a_{T-1}, s_T)$ is defined as $\E_{\pi_{\theta}(a|s), s_0\sim\rho_0}[R_{s_0}(\tau_{0:T})] = \E_{\pi_{\theta}(a|s), s_0\sim\rho_0}[\sum_{t=0}^T\gamma^tr(s_t,a_t)]$, with $a_t\sim\pi_\theta(a_t|s_t)$, and $s_{t+1} \sim P(s_{t+1}|s_t,a_t)$.
In out work, we update $\pi_\theta$ using proximal policy optimization (PPO)~\citep{schulman2017proximal} and train the agent with different initial state distribution $\rho_i$ at every training iteration $i$ as previous work~\citep{florensa2017reverse}.

\subsection{RCG for RL}
\label{subsec:rcg}
RCG for RL~\citep{florensa2017reverse} is a method that automatically adapts the initial state distribution to efficiently optimize goal-oriented tasks.
For robotic manipulation tasks with sparse rewards, standard RL methods perform poorly since useful rewards are seldom seen. 
To solve this problem, RCG adapts the current initial state distribution $\rho_{i-1}$ to a new one $\rho_{i}$ at each training iteration $i$. The initial state $s_0$ for an episode is then sampled from $\rho_i$.

At the beginning, some \emph{good starts} are generated from the states nearby the goal state $s^g$.
They will be added to a good starts pool and become the first source to sample more possible initial states, which is the first initial state distribution $\rho_1$. 
After one training iteration, the initial states $s_0 \sim \rho_1$ that satisfy the condition $R_{min} < R(\pi_1, s_0) < R_{max}$ are recognized as \emph{good starts} and added to the good starts pool. 
$R(\pi_1, s_0)$ is the expected return of $s_0$ and is calculated based on the rewards and the values outputted by the critic.
The good starts, together with some states sampled from the previous good starts pool, will be selected to sample initial states for the next training iteration.
That is, the second initial state distribution $\rho_2$ are based on these selected states.
After another training iteration, the initial states $s_0 \sim \rho_2$ will follow the processes mentioned above.
These processes will continue until the training ends. 
It should be noted that because $R_{min}$ and $R_{max}$ in the condition can be interpreted as the bounds on the probability of success, a \emph{good start} in RCG is an initial state with certain possibility to succeed.

The method that samples more initial states for the next iteration is called the \emph{sampling nearby process} (\textbf{Procedure} \ref{proc:sample_nearby}). 
This process applies noise to action space to generate a new feasible state $s'$ from a seed state $s$. 
Due to this process, the state space that the agent has mastered will gradually expand, and it is more likely for a distant initial state to reach the goal.

Three assumptions are introduced to apply RCG to a wide range of learning problems~\citep{florensa2017reverse}. These assumptions include: (1) The agent can be reset to any start state at the beginning of all trajectories. (2) At least one goal state is provided. (3) All start states generated by taking uniformly sampled random actions have a non-zero possibility of reaching the given goal state.
The first assumption allows us to reset the initial state to any state sampled from the good starts pool, and the second assumption guarantees that at least one goal state is given.
Nonetheless, this does not mean that we have a policy to find states that are arbitrarily close to the goal.
We are only able to focus on the initial states around the goal state at the beginning of learning.
Additionally, an agent can sometimes but not always get rewards from these initial states, thus achieving stronger learning signals~\citep{florensa2017reverse}.

\subsection{Generative Adversarial Parallelization (GAP)}
\label{subsec:gap}
\citet{pfau2016connecting} have pointed out the connections between GAN and AC methods in RL as well as several stabilizing strategies that might be beneficial for each class of models.
Among the strategies being proposed for stabilizing GAN training, GAP~\citep{im2016generative} is the one with potential to benefit RCG for AC methods. 
GAP eliminates the tight coupling between a generator and discriminator by simultaneously training $m$ adversarial models and exchanging their discriminators every $K$ iterations. 
The hyperparameter $m$ is the number of models trained in parallel and $K$ is the exchange rate.
The experiments show that this method can improve mode coverage, convergence and quality of the model.

\section{Methodology}
\label{sec:method}

In this section we first introduce some ideas of improving RCG and how the parallelized approach proposed for GAP, as an operator, can put these ideas into practice. 
Our detailed algorithm is shown in Section \ref{subsec:algorithms}.

By the sampling nearby process, theoretically the agent can learn to reach the goal from the whole feasible state space, but it might take a long time to learn a general policy.
Based on the algorithm, at the first training iteration, the initial state distribution $\rho_1$ will contain part of the good initial states and the states nearby them. 
Among these states $s_0$, only some of them will satisfy the condition $R_{min} < R(\pi_1, s_0) < R_{max}$, be stored in the good start pool, and run the sampling nearby process for the next iteration.
Therefore, the \emph{new states} contained in the next initial state distribution $\rho_2$ but not seen before (not yet added to the good start pool) might not be uniformly distributed.

Although at training iteration $i$, the initial state distribution $\rho_i$ will contain part of the good start pool to prevent unlearning, due to the mechanism mentioned above, $\rho_i$ might still be biased. 
If the policy only successfully learns from the initial states in some specific region at the beginning, there will be two possible results. 
(1) This process becomes a vicious cycle. The good start pool will become biased, and the policy will seriously overfit. 
(2) Since the sampling nearby process only generate the unlearned initial states occasionally, the expansion rate becomes very slow, so does the speed of convergence.

\subsection{Improvement of RCG for RL}
\label{subsec:improve_rcg}

\begin{figure*}[b!]
    \centering
    \subfigure[Grasping Task]{
        \label{fig:env_grasp}
        \includegraphics[height=0.3\linewidth]{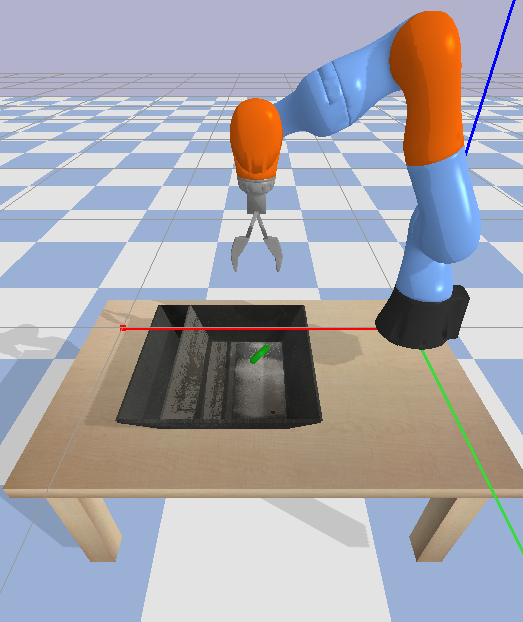}}%
    \subfigure[Door Opening Task]{
        \label{fig:env_door}
        \includegraphics[height=0.3\linewidth]{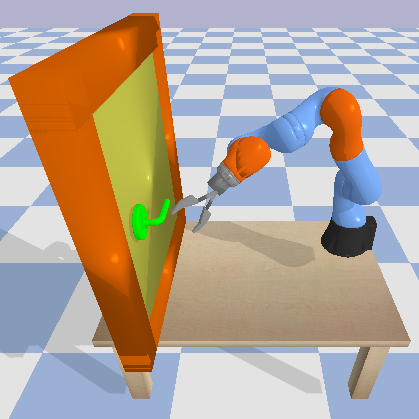}}
    \subfigure[Key Insertion Task]{
        \label{fig:env_key}
        \includegraphics[height=0.3\linewidth]{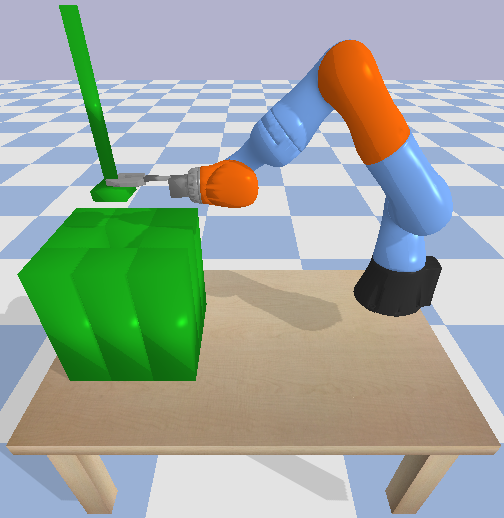}}
    \captionof{figure}{Robotic Manipulation Environments}
    \vspace{-15pt}
    \label{fig:env}
\end{figure*}

\begin{figure*}[t!]
    \centering
    \subfigure[Model 1]{
        \label{fig:RCG_good_pos1}
        \includegraphics[height=0.3\linewidth]{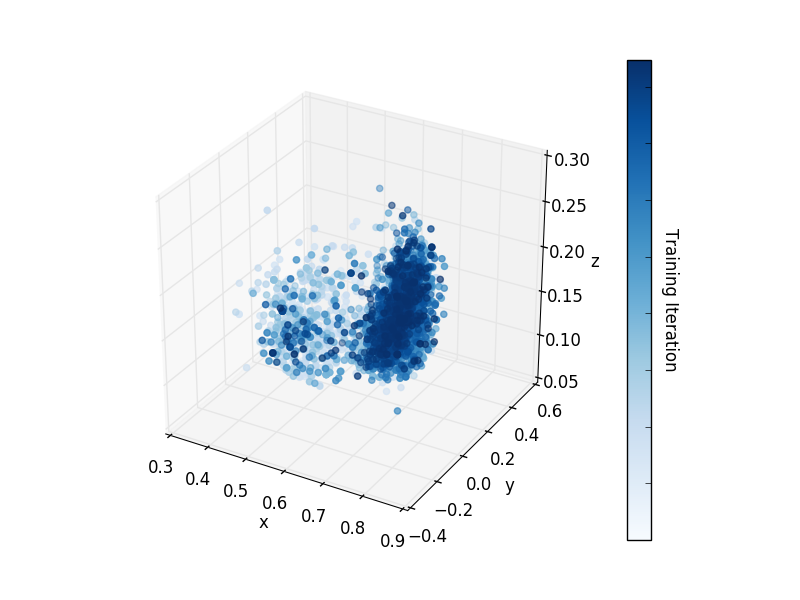}}%
    \subfigure[Model 2]{
        \label{fig:RCG_good_pos2}
        \includegraphics[height=0.3\linewidth]{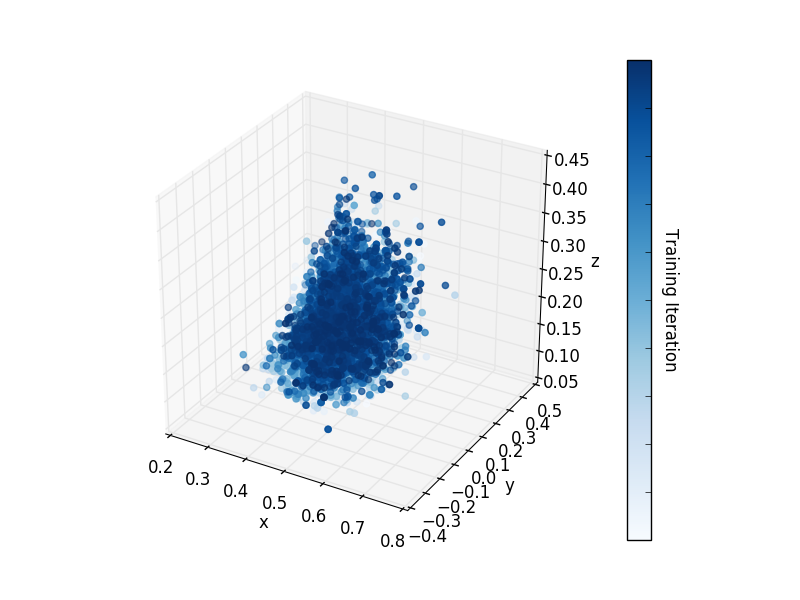}}
    \captionof{figure}{Learned initial positions of the grasping task}
    \vspace{-15pt}
    \label{fig:good_pos_RCG}
\end{figure*}

We consider an environment where a seven DOF robotic arm learns how to grasp an object on a table, as shown in Fig. \ref{fig:env_grasp}. 
For this task, the good starts at the beginning will be the states of the robotic gripper close to the object. 
Fig. \ref{fig:good_pos_RCG} shows the initial positions that the robotic gripper can successfully grasp the object after training with RCG.  
The colors in the figure represent the order of these initial positions being learned. 
The object might appear at $0.45 \leq x \leq 0.55$, $-0.05 \leq y \leq 0.05$, and $z = -0.1$.
Fig. \ref{fig:RCG_good_pos1} and Fig. \ref{fig:RCG_good_pos2} show the learned initial positions of two models trained by the same method with different initialization, and the gripper at these positions can successfully grasp the object.
From Fig. \ref{fig:good_pos_RCG} we can observe that the distribution of learned initial positions is nonuniform, and the learned initial positions might be very different even if the models are trained by the same method.

Ideally, the agent should learn from easier states (shorter distance between the gripper and the object) to harder states (longer distance between the gripper and the object). 
However, since the sampling nearby process usually only generate states that cover part of the easier state space, the agent might start to learn from further states without generally mastering most of the easier states. 
That is, for the easier state space $S_e^0$ that the agent thinks to have learned (all the states $s_0$ in $S_e^0$ satisfy $R_{max} \leq R(\pi_i, s_0)$) actually contains states that has not yet been mastered ($R(\pi_i, s_0) < R_{max}$). 
This phenomenon can especially be found in Fig. \ref{fig:RCG_good_pos1}.  The sparse distribution of the initial positions at $z \leq 0.1$ indicates that the agent does not learn the easier states well.
Moreover, the sampling nearby process makes the expansion start from only part of the easier states, so it can be observed that the well-learned initial positions accumulate mostly at $0.6 \leq x \leq 0.7$.
Finally, it becomes extremely difficult for the agent to learn a general policy since the good start pool is already biased, and the policy only overfits to the states it has seen before.

The issue can be less serious. 
Even if the good start pool tends to be biased, the random mechanism in sampling nearby process can still occasionally generate the unlearned easier states for the agent.
Therefore, the agent can prevent the occurrence of seriously overfitting.
However, the speed of convergence will be very slow due to tardy expansion rate.
This phenomenon can be found in Fig. \ref{fig:RCG_good_pos2}. 
Although the learned initial positions tend to accumulate at $0.1 \leq y \leq 0.3$, the sampling nearby process can sometimes generate the initial positions at $-0.1 \leq y \leq 0.1$ for the agent to learn.
Nonetheless, since most of the learned initial positions in good start pool are at $0.1 \leq y \leq 0.3$, relatively less initial positions at $-0.1 \leq y \leq 0.1$ will be generated for each training iteration.
As a result, the expansion rate at $-0.1 \leq y \leq 0.1$ becomes tardy, and the agent learns a general policy at a slow pace, which leads to poor convergence of the model.

\subsection{Parallelized Reverse Curriculum Generation for AC Methods}
\label{subsec:algorithms}

In Section \ref{sec:motivation}, we list the similarities between GANs and RCG for AC methods. 
Since the cause of mode collapse in GANs is similar to the cause of overfitting in RCG for AC methods, we try to apply the approaches of solving mode collapse to RCG.
Among these approaches, GAP seems to be a suitable reference for us. 
The reason is that in GAP, each generator needs to generate samples from different modes to successfully fool all the discriminators, so the mode coverage can thus be improved~\citep{im2016generative}.
In our case, an actor can learn a more general policy if it can adapt to different critics with distinct learned initial positions (e.g., Fig. \ref{fig:RCG_good_pos1} and Fig. \ref{fig:RCG_good_pos2}).
Moreover, since different critics can guide an actor to learn from more unseen initial states, the exploration of the model can be faster.
Therefore, we propose the method that utilizes the parallelized approach in GAP to improve the performance and convergence of RCG for AC methods.

\begin{algorithm}[t!]
  \SetAlgoLined
  \KwIn{(1) number of models $m$, (2) number of training iterations $I$, (3) goal states $s^g$, (4) number of states sampled by Procedure \ref{proc:sample_nearby} $N_{new}$, (5) number of states sampled from a good start pool $N_{old}$, (6) boundary number of filtering initial states $R_{min}$ and $R_{max}$, (7) exchange rate $K$}
  \KwOut{policy $\pi^*$}
    Let $M=\{(A_1,C_1),\cdots,(A_m,C_m)\}$.\\
    $starts_{old} \leftarrow s^g;$\\
    $starts \leftarrow s^g$;\\
    \For{ $iteration \leftarrow 1$ to $I$ } {
        \For{ $n \leftarrow 1$ to $m$ } {
            $starts[n] \leftarrow SampleNearby(starts[n], N_{new})$;\\
            $starts[n].append(sample(starts_{old}[n], N_{old}))$;\\
            $\rho \leftarrow Unif(starts[n])$;\\
            Run policy $A_{n}$ with start state distribution $\rho$.\\
            Compute the expected rewards $rews$.\\
            $starts[n] \leftarrow select(starts[n], rews, R_{min}, R_{max})$;\\
            $starts_{old}[n].append(starts[n])$;
        }
        \For{ $n \leftarrow 1$ to $m$ } {
            Optimize the objective functions w.r.t. $A_n, C_n$.
        }
        \If{ $iteration \% K == 0$ } {
            Randomly generate $\frac{m}{2}$ pairs (include all models) with indices $(i,j)$ for replacement.\\
            Swap $C_i$ and $C_j$.
        }
    }
  \caption{Training procedure of PRCG}
  \label{alg:prcg}
\end{algorithm}

\begin{procedure}[t!]
  \caption{SampleNearby() (from \citep{florensa2017reverse})}
  \label{proc:sample_nearby}
  \KwIn{(1) initial state pool $starts$, (2) the number of running the expansion process $N_{total}$, (3) the standard deviation for the expansion process $\Sigma$, (4) RL environment $env$, (5) the number of the outputted initial states $N_{new}$}
  \KwOut{expanded initial state pool $starts_{new}$}
    \While{$len(starts) < N_{total}$}{
        $s \sim Unif(starts)$;\\
	    \While{not terminated}{
	        $a \sim \mathcal{N}(0, \Sigma)$;\\
	        $s', t \leftarrow env.step(s, a)$\\
	        $starts.append(s')$;\\
	        $s \leftarrow s'$
	    }
    }
    $starts_{new} \leftarrow sample(starts, N_{new})$;
\end{procedure}

We propose a method that uses RCG to simultaneously train $m$ AC models.
By randomly swapping their critics every $K$ iterations, this method can avoid the poor generality and exploration of learning caused by the issues mentioned in Section \ref{subsec:improve_rcg}. 
We call the proposed method \emph{Parallelized Reverse Curriculum Generation} (PRCG), and the final policy is chosen from the one that performs best in the environment among the $m$ models. 
The detailed algorithm of our method is shown in Algorithm \ref{alg:prcg}. 
As mentioned in Section \ref{sec:motivation}, a critic in RCG serves as a filter of the initial states and will choose the initial states for the next training iteration. 
This mechanism is done when computing the expected rewards $rews$ in the algorithm.

It should be noted that the parallelized approach in PRCG can be applied to a large state space, e.g., RGB states, without losing too much exploration efficiency.
No matter how large a state space is, the possible number of initial states is still decided by the action space since all initial states are generated by applying actions to a given state (Procedure \ref{proc:sample_nearby}).
That is, the agent does not need to learn from all possible states before reaching the target one, and the learned initial state space only depends on the action space. 
Therefore, scaling to a large state space will not seriously affect the efficiency of the algorithm.

\subsection{Different Training Strategies for PRCG}
\label{subsec:diff_prcg}

In PRCG, the critics are swapped to share good learning signals and environmental information. 
However, to achieve this goal, swapping critics does not seem to be the only solution. 
For example, swapping initial state pools ($starts_{old}$ and $starts$ in Algorithm \ref{alg:prcg}) and using a common initial pool might also work. 
In this section, we analyze the effects of different training strategies for PRCG and find that swapping critics leads to faster exploration and maintains stable learning.

\begin{figure*}[t!]
    \centering
    \subfigure[No Swapping]{
        \label{fig:no_swapping}
        \includegraphics[width=0.3\linewidth]{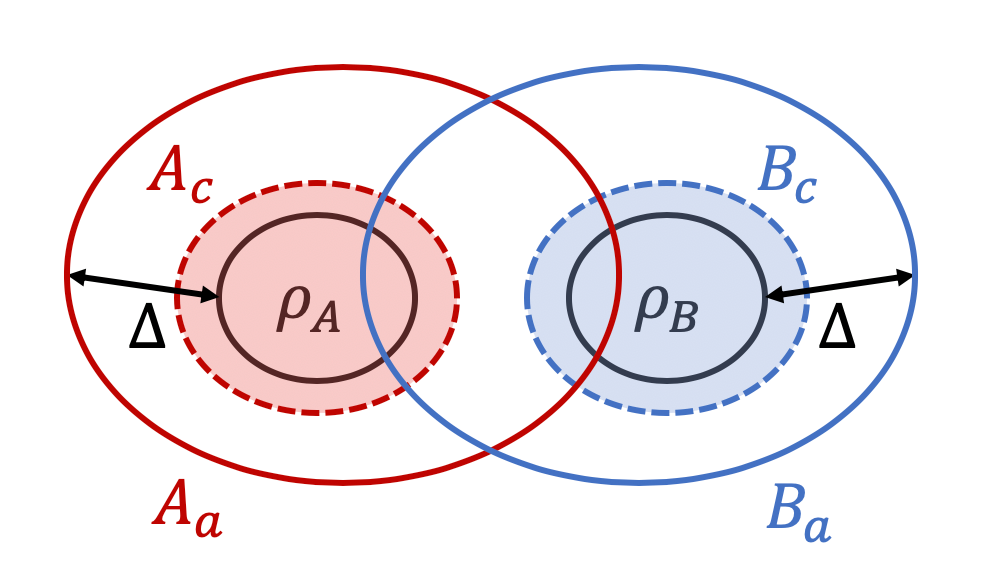}}
    \subfigure[Swap Critics]{
        \label{fig:swap_critics}
        \includegraphics[width=0.3\linewidth]{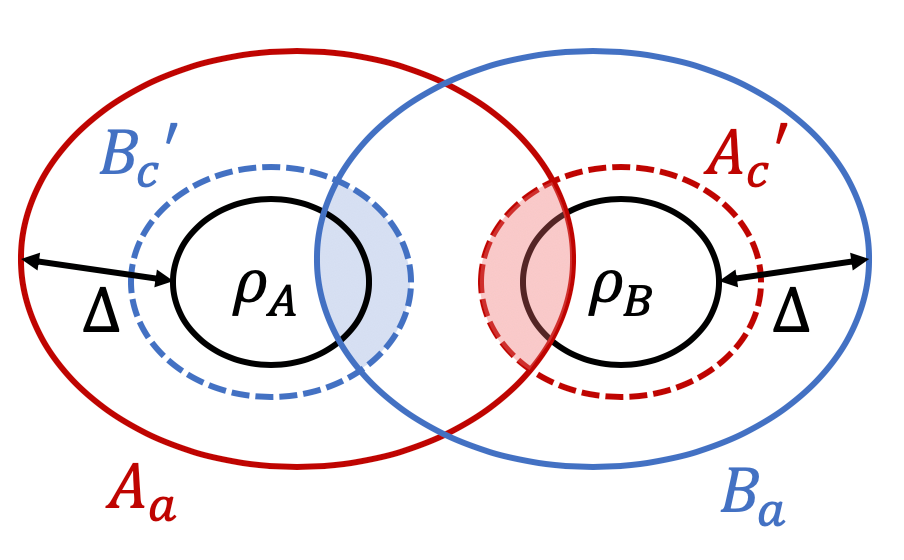}}
    \subfigure[Swap Initial State Pools]{
        \label{fig:swap_inits}
        \includegraphics[width=0.3\linewidth]{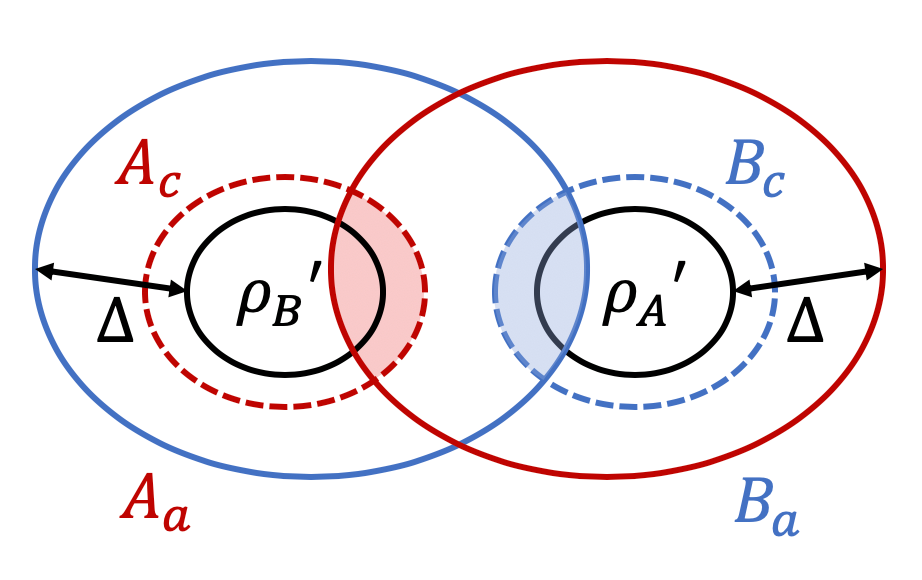}}
    \captionof{figure}{Comparison between swapping critics and swapping initial state pools in PRCG}
    \vspace{-15pt}
    \label{fig:diff_swapping}
\end{figure*}

We first compare the difference between swapping critics and swapping initial state pools. 
Fig. \ref{fig:diff_swapping} demonstrates the effects of different swapping strategies. 
We consider that PRCG is trained with two models \textcolor{red}{$A$ (red)} and \textcolor{blue}{$B$ (blue)}. 
In Fig. \ref{fig:diff_swapping}, $\rho_A$ and $\rho_B$ are the initial state pools for $A$ and $B$ respectively at the beginning of a training iteration. 
After the sampling nearby process, $\rho_A$ and $\rho_B$ will be expanded to $\rho_A+\Delta$ and $\rho_B+\Delta$ respectively, which are the red and blue solid-line ovals in the figure. 
The critics of $A$ and $B$ are represented as $A_c$ and $B_c$. 
Their filtering effects are shown by the red and blue dashed-line ovals in the figure. 
Only the initial states within these dashed ovals will be chosen as the good starts for the next training iteration. 
The actors of $A$ and $B$ are represented as $A_a$ and $B_a$.

The red and blue shaded areas in Fig. \ref{fig:diff_swapping} show the good starts chosen for $A$ and $B$ respectively. 
For each model, the good starts will lie in the area of the dash-line oval that intersects with the solid-line oval in the same color. 
In Fig. \ref{fig:swap_critics}, the good starts selected for $A_a$ and $B_a$ are further from the original ones ($\rho_A$ and $\rho_B$) while in Fig. \ref{fig:swap_inits}, the good starts selected for $A_a$ and $B_a$ are still close to the the original ones. 
As a result, swapping critics can explore more than swapping initial states. 
In our experiments, we will demonstrate that swapping critics can achieve faster exploration in all tasks.

Compared with maintaining separate initial state pools, using a common initial state pool might make the training unstable. 
If the good starts collected by each model are in separate regions, then each actor might be very unfamiliar with some initial states sampled in the next training iteration. 
This will hurt the performance of the models. 
In our experiments, we will show that maintaining separate initial state pools performs better than using a common one in general.

\section{Experimental Results}
\label{sec:result}

\subsection{Experimental Setup}
We built three simulation environments (Fig. \ref{fig:env}) using \textit{PyBullet}~\citep{coumans2018}. These environments simulate a 7-DOF KUKA\footnote{https://www.kuka.com/} robot arm doing three different tasks:  grasping a cuboid, opening a door, and inserting a key into a keyhole.
Following are the detailed settings of these tasks:

\noindent
\begin{itemize}[leftmargin=*]
\item \textbf{State}: A state consists of the 7 joint angles of the robot arm, the position as well as Euler orientation of the end effector, and the relative position as well as Euler orientation between the end effector and the target object. 
The target object in the grasping task is the cuboid, in the door opening task is the door knob, and in the key insertion task is the keyhole.
\item \textbf{Action}: An action is to apply 7 joint angle variations to the current joint angles. The dimension is 7, and the range of each variation is [-0.2, 0.2] rad.
All three tasks use the same action setting.
Note that here we use the variation of joint angles as the action and do not fix the direction of the gripper to be down, so the configuration of the robot arm becomes more diverse and make the learning problem more complex.
\item \textbf{Reward}: In the grasping task, a reward of 1 will only be given when the object is picked up;
in the door opening task, a reward of 1 will only be given when the door is pulled open;
in the key insertion task, a reward of 1 will only be given when the key is completely inserted into the keyhole.
Otherwise, the reward will be 0.
\item \textbf{Termination Conditions}: The maximum number of steps $T_{max}$ in one episode is 10, and an episode would terminate if one of the following conditions is satisfied: 
(1) The step number reaches $T_{max}$. 
(2) In the grasping task, the gripper is low enough to try to grasp the cuboid;
in the door opening task, the gripper is close enough to the door to try to pull the doorknob;
in the key insertion task, the gripper is low enough to try to insert the key into the keyhole.
\end{itemize}
The source code of our simulation environments has been uploaded to GitHub: https://github.com/SarahChiu/RobotEnv-PyBullet.

The actor and critic for each model are both represented as a (256, 256) multi-layer perceptron (MLP) with a tangent hyperbolic activation function at the first layer and a linear activation function at the second layer. 
The models with Gaussian policies and state-value function approximators are trained with the actor-critic style PPO algorithm~\citep{schulman2017proximal} as implemented in OpenAI Baselines~\citep{baselines}.
We train 500 iterations for the grasping task, and 200 iterations for the door opening as well as key insertion tasks.
We set the optimizing step size to $3 \times 10^{-4}$, the batch size to 2056, and the discount factor $\gamma$ to $0.99$. 
For the setting of RCG, most hyperparameters are the same as proposed in the original paper~\citep{florensa2017reverse}, including $M$, $N_{old}$, and $N_{new}$.
Since our maximum number of steps in one episode is smaller, we set our $R_{min}$ and $R_{max}$ to be 0.93 and 0.96.
These values imply the bounds of the probability of success $R$ for an initial state to reach the goal at some time step $T$, and $R$ is calculated by $\E_{\pi_{\theta}(a|s)}[\sum_{t=0}^T\gamma^tr(s_t,a_t)]$.

\subsection{Results}

\begin{figure*}[t!]
    \centering
    \subfigure[Grasping Task]{
        \label{fig:ps_grasp_exp4}
        \includegraphics[width=0.3\linewidth]{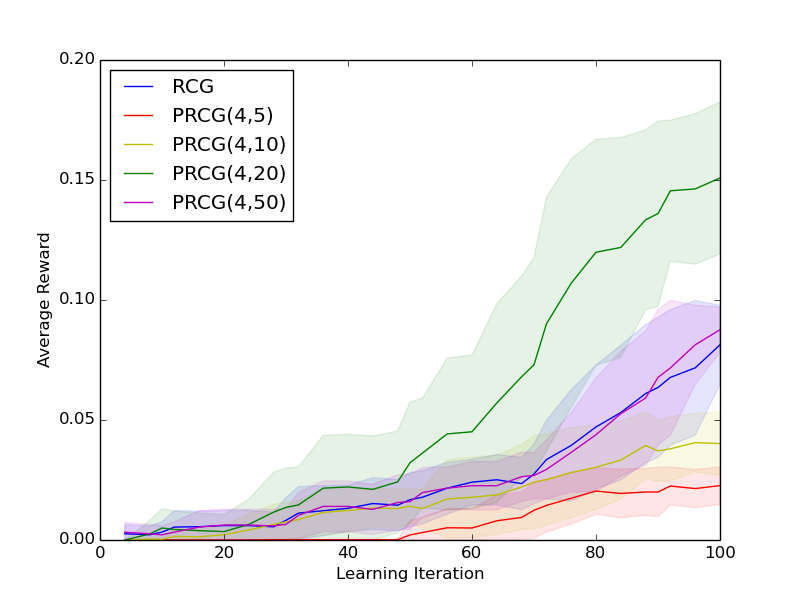}}
    \subfigure[Door Opening Task]{
        \label{fig:ps_door_exp4}
        \includegraphics[width=0.3\linewidth]{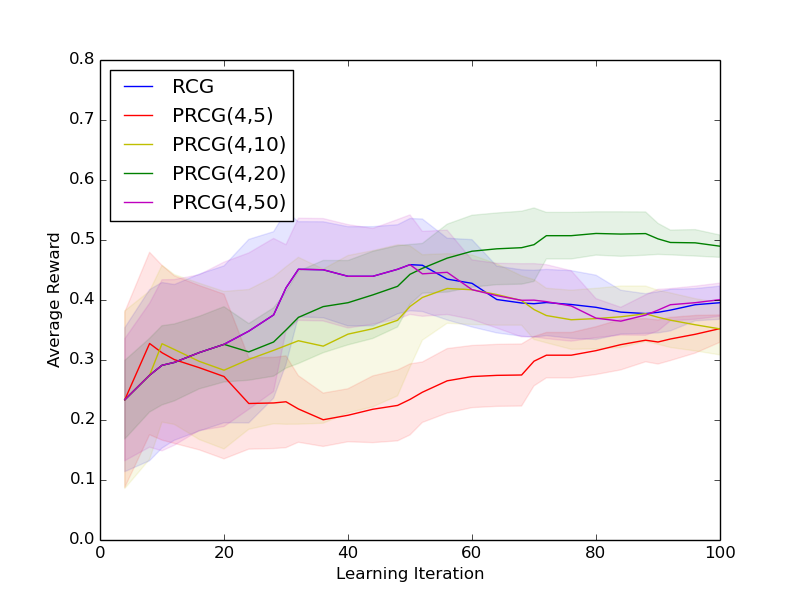}}
    \subfigure[Key Insertion Task]{
        \label{fig:ps_key_exp4}
        \includegraphics[width=0.3\linewidth]{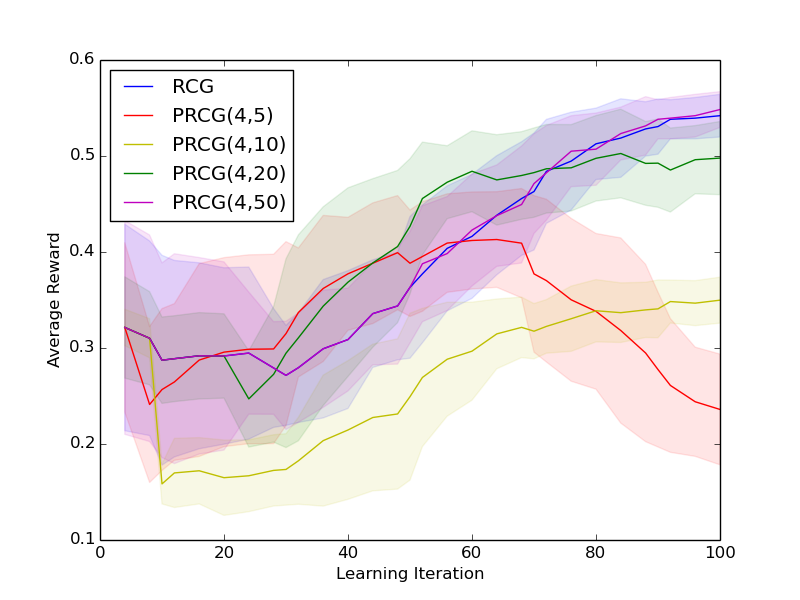}}
    \captionof{figure}{Parameter Search Results for Experiment 4 (Near, V)}
    \vspace{-15pt}
    \label{fig:ps_exp4}
\end{figure*}

\begin{figure*}[t!]
    \centering
    \subfigure[Grasping Task]{
        \label{fig:ps_grasp_exp5}
        \includegraphics[width=0.3\linewidth]{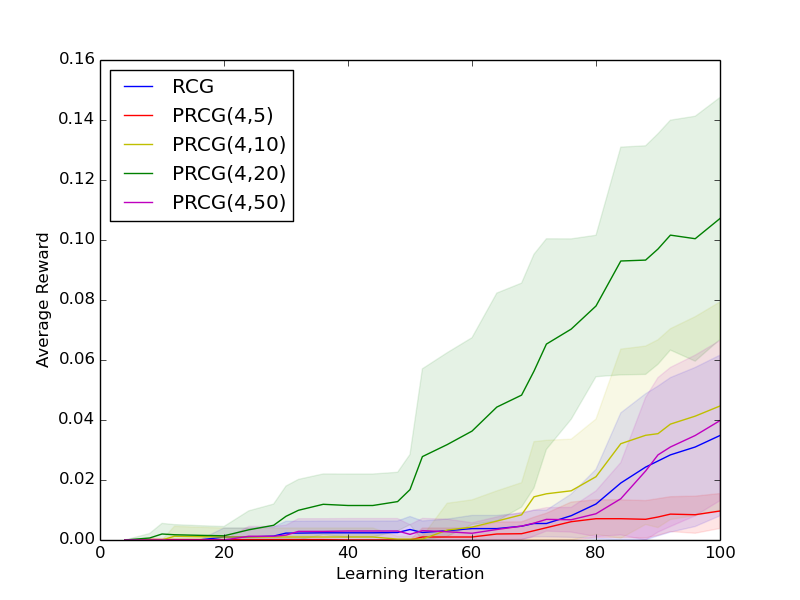}}
    \subfigure[Door Opening Task]{
        \label{fig:ps_door_exp5}
        \includegraphics[width=0.3\linewidth]{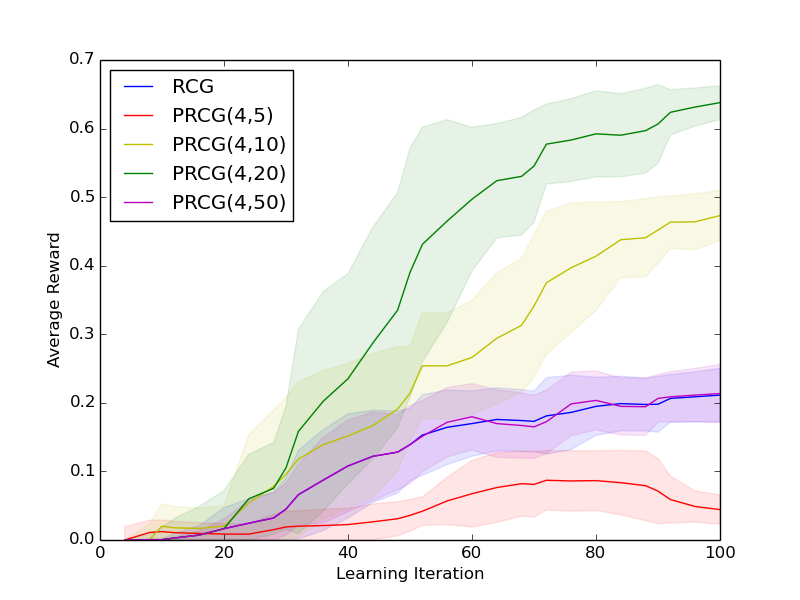}}
    \subfigure[Key Insertion Task]{
        \label{fig:ps_key_exp5}
        \includegraphics[width=0.3\linewidth]{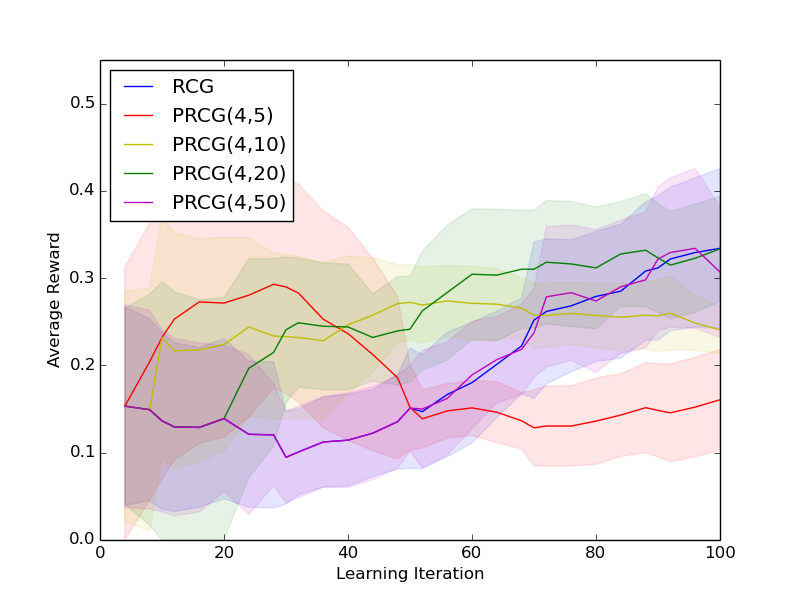}}
    \captionof{figure}{Parameter Search Results for Experiment 5 (Mid, V)}
    \vspace{-15pt}
    \label{fig:ps_exp5}
\end{figure*}

In our experiments, we will show the effects of PRCG by answering the following three questions:

\noindent
\begin{itemize}[leftmargin=*]
\item $\text{Q}_1$: Does PRCG make the learned initial positions more uniformly distributed?
\item $\text{Q}_2$: Does PRCG generally improve the performance of the policy?
\item $\text{Q}_3$: What are the effects of different training strategies for PRCG?
\item $\text{Q}_4$: Can the parallelized approach in PRCG be used to improve other AC based RL algorithms?
\end{itemize}

To test the general performance of each model, we evaluate the trained policies from three different levels and two different kinds of initial poses of the gripper. 
The three levels are \textit{near}, \textit{middle}, and \textit{far}, which represent the initial distance between the gripper and the target object. The two different initial poses are \textit{fixed} and \textit{variable}, which represent if the initial position of the gripper varies.
In real-life robotic applications, the initial pose of a robot arm is usually fixed. However, while performing a task, the pose of the robot arm might become any feasible one. Therefore, it is also important to test the models with variable initial poses.
There will be total six ($3 \times 2$) different experiments to evaluate the policies, and these experiments are summarized in Table \ref{table:experiment}.

\begin{table}[h!]
\begin{center}
\begin{tabular}{ |c||c|c|c| }
    \hline
     & Near & Middle (Mid) & Far \\
    \hline\hline
    Fixed (F) & Experiment 1 & Experiment 2 & Experiment 3\\
    \hline
    Variable (V) & Experiment 4 & Experiment 5 & Experiment 6\\
    \hline
\end{tabular}
\end{center}
\captionof{table}{Testing Experiments}
\vspace{-10pt}
\label{table:experiment}
\end{table}

In our proposed algorithm, the hyperparameters, $m$ and  $K$, need to be properly chosen to reach the desired performance.
PRCG is trying to produce a synergistic effect among different AC pairs, so we can expect a better performance for larger $m$.
Based on the computational resources we have, we set $m$ to be 2 and 4 in our experiments.

$K$ is not very selective.
The principle of choosing $K$ is that $K$ cannot be too small (e.g. $K < 5$)  or too large (e.g. $K > 30$).
The values of $K$ that are too small will make training unstable; the values of $K$ that are too large will make the improvement insignificant.
We found that $K$ around 20 performs well in our experiments.

Here we show the results of parameter search done on the three tasks, and we will use the best $K$ in our later experiments.
For each $K$, the model is trained for 100 iterations.
We use PRCG($m$,$k$) to represent that PRCG is trained with $m$ models and $K = k$.
All experiments were repeated 4 times with random initializations.
Fig. \ref{fig:ps_exp4} and Fig. \ref{fig:ps_exp5} show the results of RCG and PRCG with $K = 5$, $10$, $20$, $50$ for Experiment 4 and 5.
It can be observed that when $K = 5$, PRCG makes the training unstable; when $K = 10$, PRCG outperforms RCG in some experiments; when $K = 20$, PRCG outperforms RCG in most experiments; when $K = 50$, PRCG and RCG have almost the same result.
This shows that exchanging critics plays an important role on improving the performance. 
Since $K = 20$ seems to have a better performance, we use this setting in the later experiments.

\subsubsection{Results of the learned initial position distribution}

\begin{figure*}[t!]
    \centering
    \subfigure[RCG]{
        \label{fig:grasp_RCG_gp}
        \includegraphics[width=0.3\linewidth]{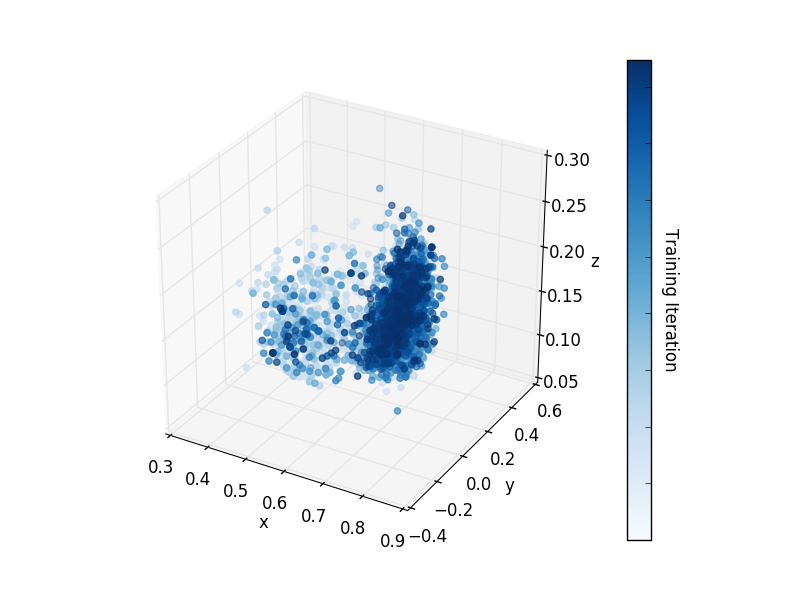}
    }
    \subfigure[PRCG(2,20)]{
        \label{fig:grasp_PRCG2_gp}
        \includegraphics[width=0.3\linewidth]{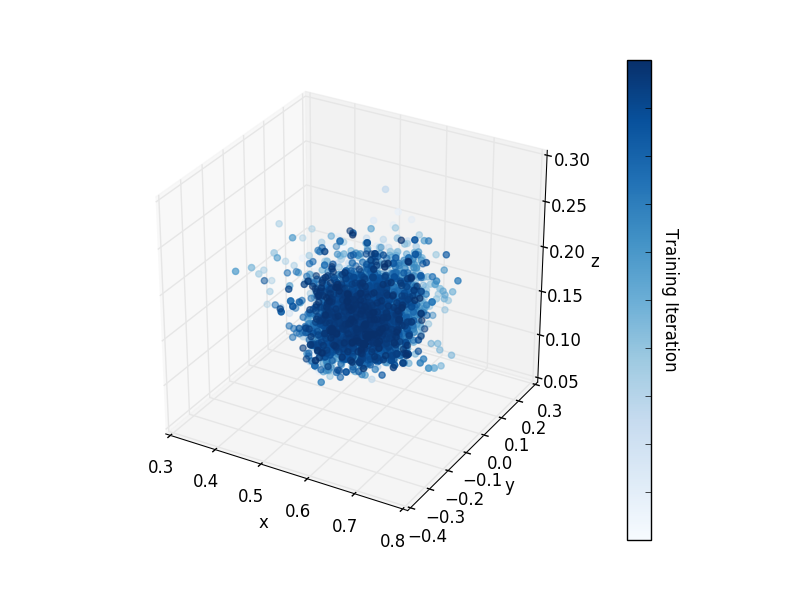}}%
    \subfigure[PRCG(4,20)]{
        \label{fig:grasp_PRCG4_gp}
        \includegraphics[width=0.3\linewidth]{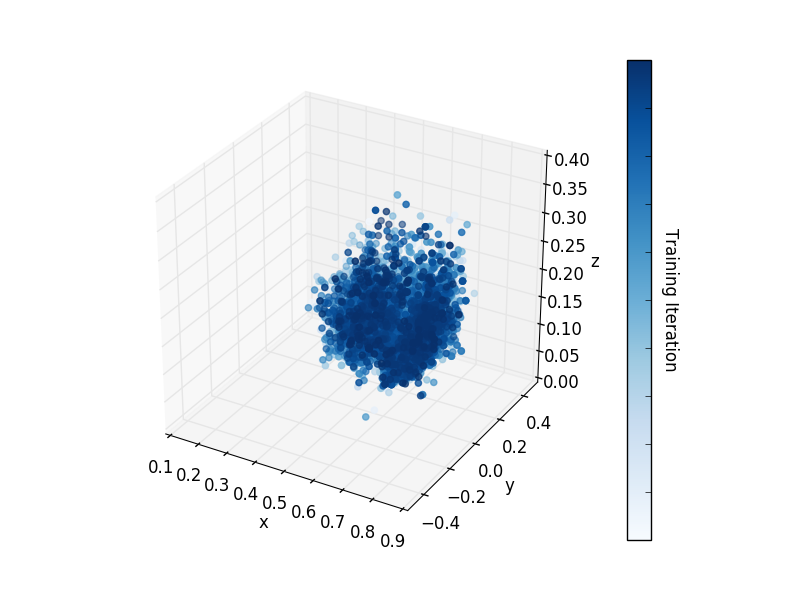}}
    \captionof{figure}{Learned initial positions of the grasping task (The target object is approximately at (0.5, 0, -0.1).)}
    \vspace{-15pt}
    \label{fig:grasp_gp}
\end{figure*}

\begin{figure*}[t!]
    \centering
    \subfigure[RCG]{
        \label{fig:door_RCG_gp}
        \includegraphics[width=0.3\linewidth]{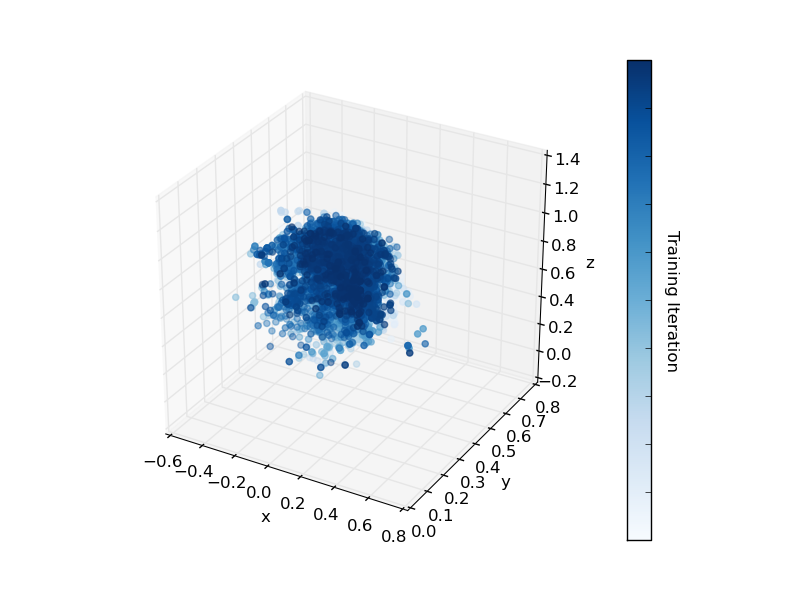}
    }
    \subfigure[PRCG(2,20)]{
        \label{fig:door_PRCG2_gp}
        \includegraphics[width=0.3\linewidth]{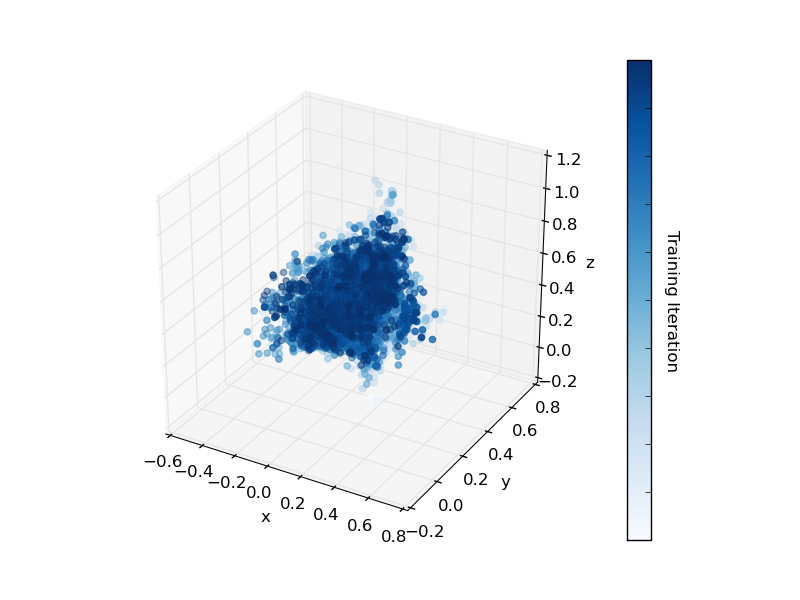}}%
    \subfigure[PRCG(4,20)]{
        \label{fig:door_PRCG4_gp}
        \includegraphics[width=0.3\linewidth]{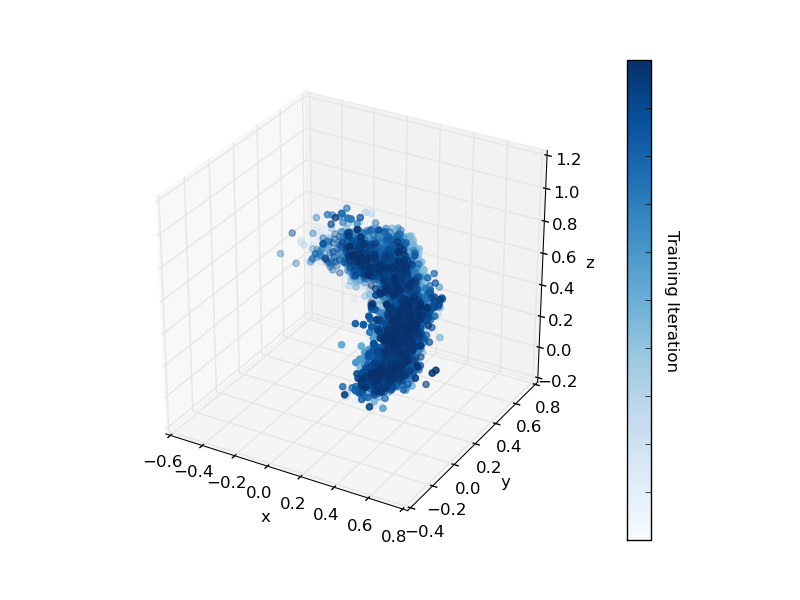}}
    \captionof{figure}{Learned initial positions of the door opening task (The target object is approximately at (0.35, 0.025, 0.2).)}
    \vspace{-15pt}
    \label{fig:door_gp}
\end{figure*}

\begin{figure*}[t!]
    \centering
    \subfigure[RCG]{
        \label{fig:key_RCG_gp}
        \includegraphics[width=0.3\linewidth]{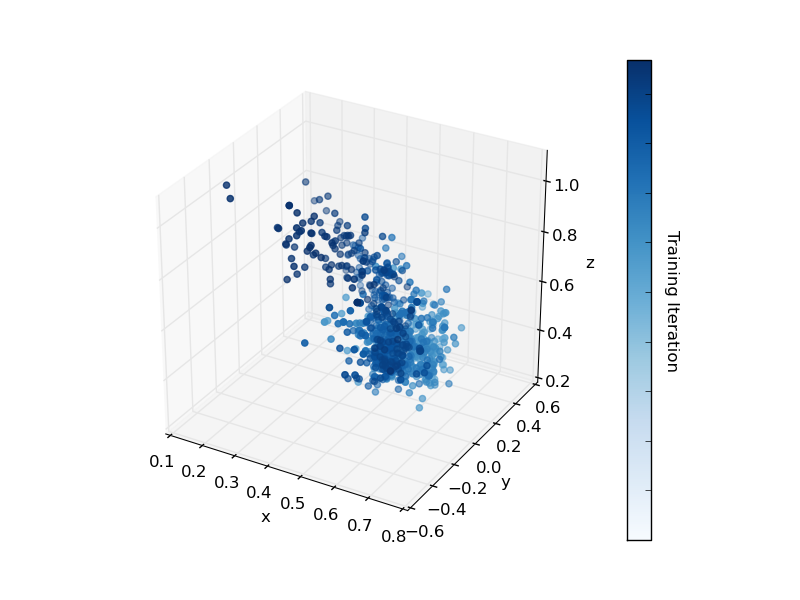}
    }
    \subfigure[PRCG(2,20)]{
        \label{fig:key_PRCG2_gp}
        \includegraphics[width=0.3\linewidth]{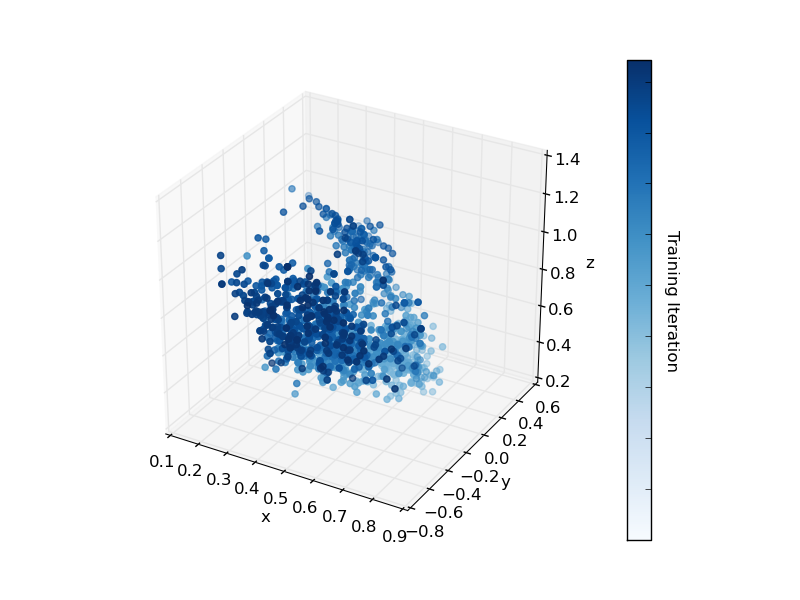}}%
    \subfigure[PRCG(4,20)]{
        \label{fig:key_PRCG4_gp}
        \includegraphics[width=0.3\linewidth]{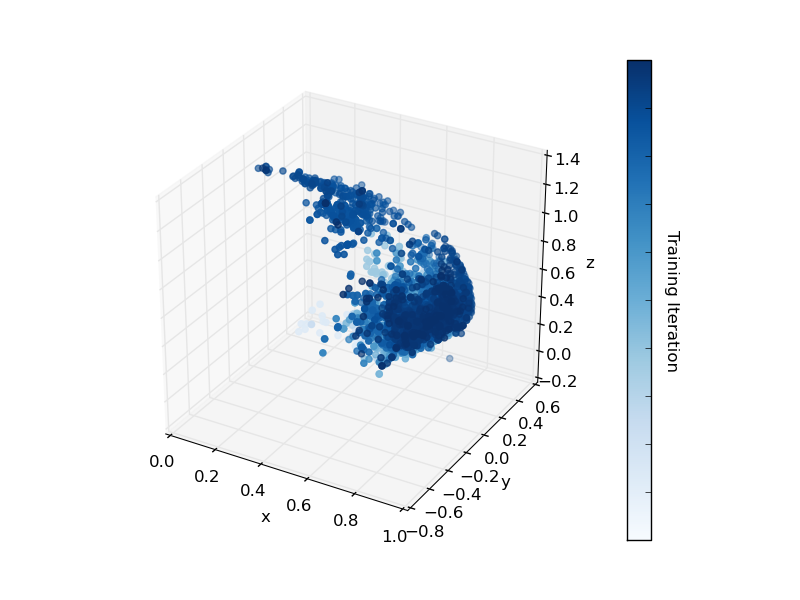}}
    \captionof{figure}{Learned initial positions of the key insertion task (The target object is approximately at (0.75, 0, 0.5).)}
    \vspace{-15pt}
    \label{fig:key_gp}
\end{figure*}

Fig. \ref{fig:grasp_gp} to Fig. \ref{fig:key_gp} show the learned initial positions of the gripper collected by the models trained with RCG, PRCG(2,20), and PRCG(4,20).
As can be seen in Fig. \ref{fig:grasp_RCG_gp} and Fig. \ref{fig:key_RCG_gp}, the distribution of the learned initial positions generated by RCG is more nonuniform, especially when the gripper is further from the target object. 
As training iteration goes by, the learned initial positions become concentrated in some undesired region (Fig. \ref{fig:grasp_RCG_gp} and Fig. \ref{fig:door_RCG_gp}) or sparse (Fig. \ref{fig:key_RCG_gp}).
In contrast, the distribution of the learned initial positions generated by PRCG(2,20) and PRCG(4,20) is more concentrated in the desired region.
It can also be observed that PRCG(4,20) has a larger expansion range of the learned initial positions than PRCG(2,20) in Fig. \ref{fig:grasp_gp}, Fig. \ref{fig:door_gp}, and Fig. \ref{fig:key_gp}. Therefore, we can expect that PRCG(4,20) will perform better when the initial state $s_0$ is further from the goal state $s_g$.

\subsubsection{General policy improvement by PRCG}

\begin{figure*}[t!]
    \centering
    \subfigure[Experiment 1 (Near, F)]{
        \label{fig:grasp_exp1}
        \includegraphics[width=0.3\linewidth]{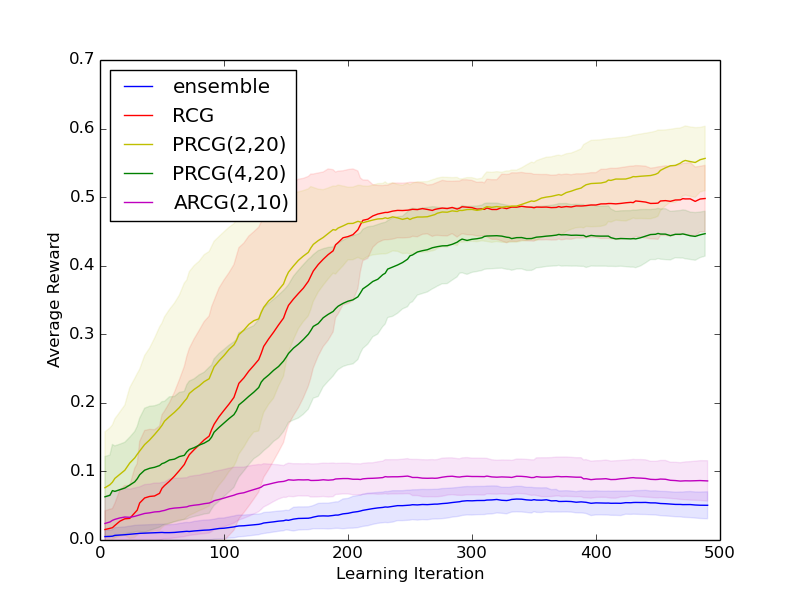}}%
    \subfigure[Experiment 2 (Mid, F)]{
        \label{fig:grasp_exp2}
        \includegraphics[width=0.3\linewidth]{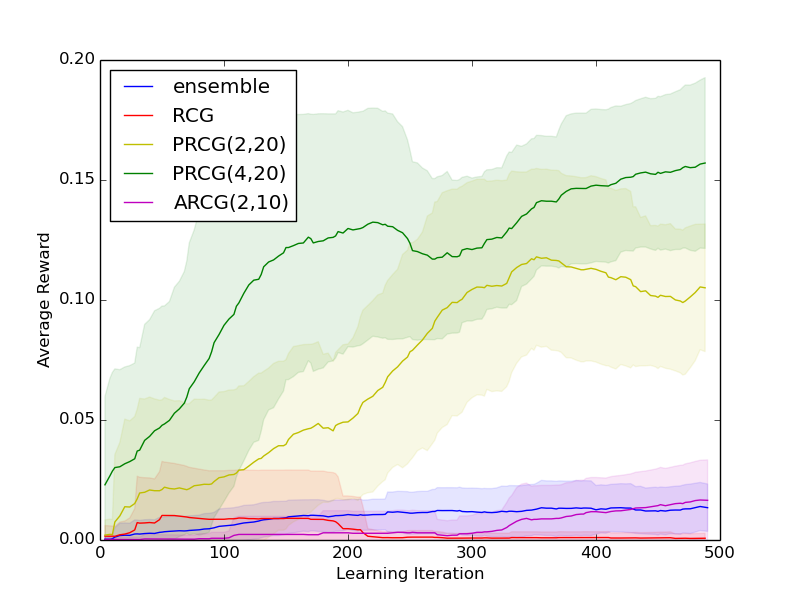}}
    \subfigure[Experiment 3 (Far, F)]{
        \label{fig:grasp_exp3}
        \includegraphics[width=0.3\linewidth]{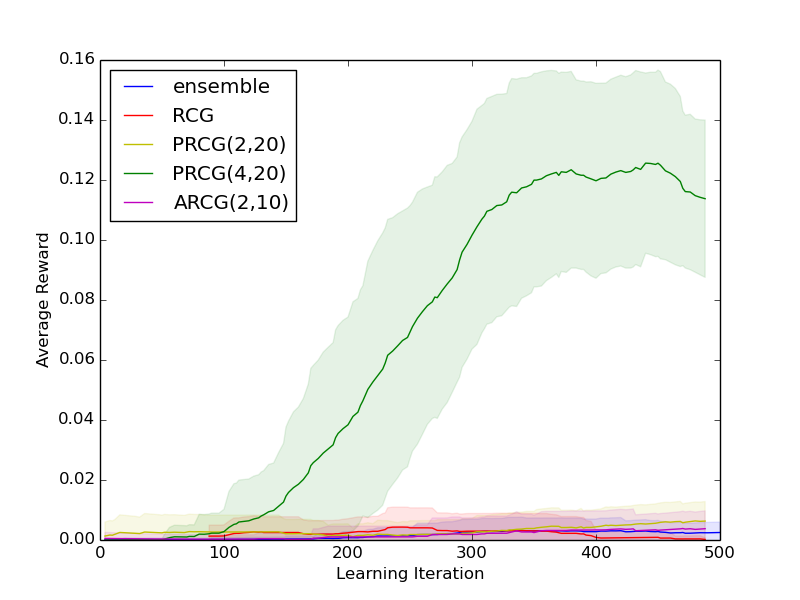}}
    \subfigure[Experiment 4 (Near, V)]{
        \label{fig:grasp_exp4}
        \includegraphics[width=0.3\linewidth]{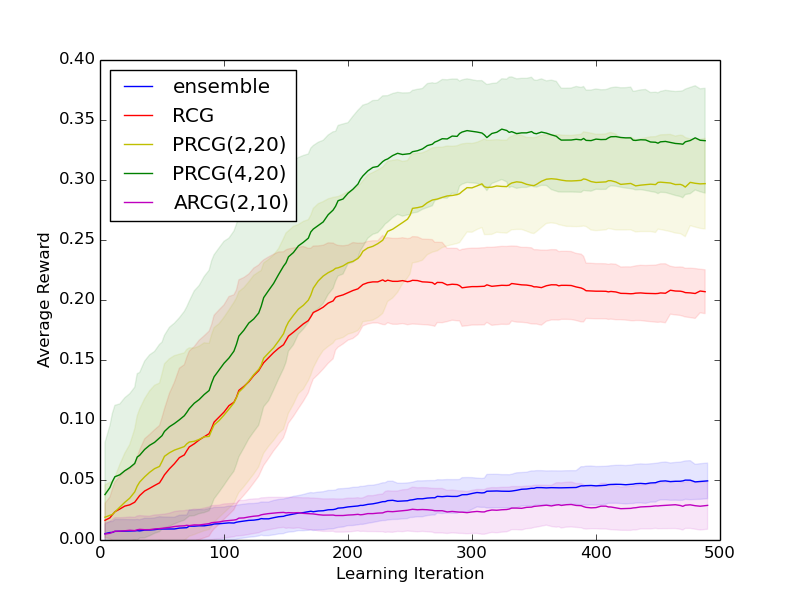}}%
    \subfigure[Experiment 5 (Mid, V)]{
        \label{fig:grasp_exp5}
        \includegraphics[width=0.3\linewidth]{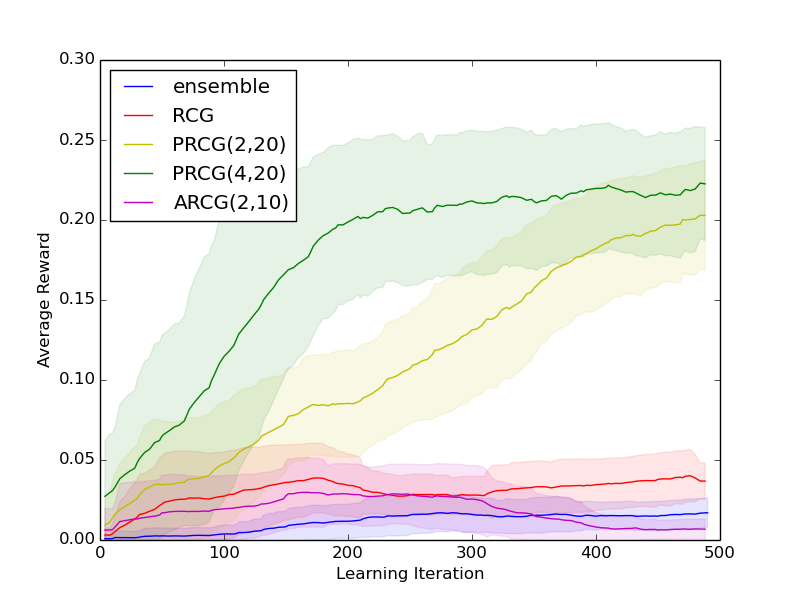}}
    \subfigure[Experiment 6 (Far, V)]{
        \label{fig:grasp_exp6}
        \includegraphics[width=0.3\linewidth]{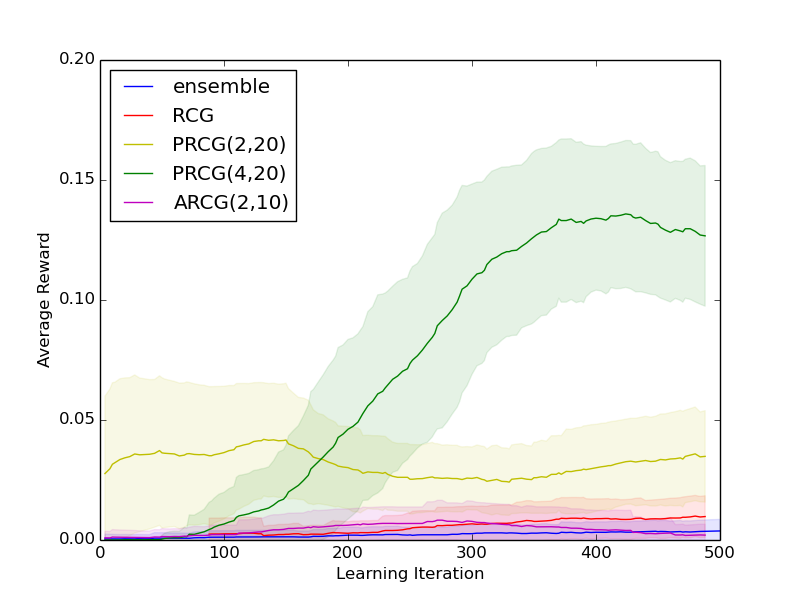}}
    \captionof{figure}{Learning curve for the grasping task}
    \vspace{-15pt}
    \label{fig:grasp_lc}
\end{figure*}

\begin{figure*}[t!]
    \centering
    \subfigure[Experiment 1 (Near, F)]{
        \label{fig:door_exp1}
        \includegraphics[width=0.3\linewidth]{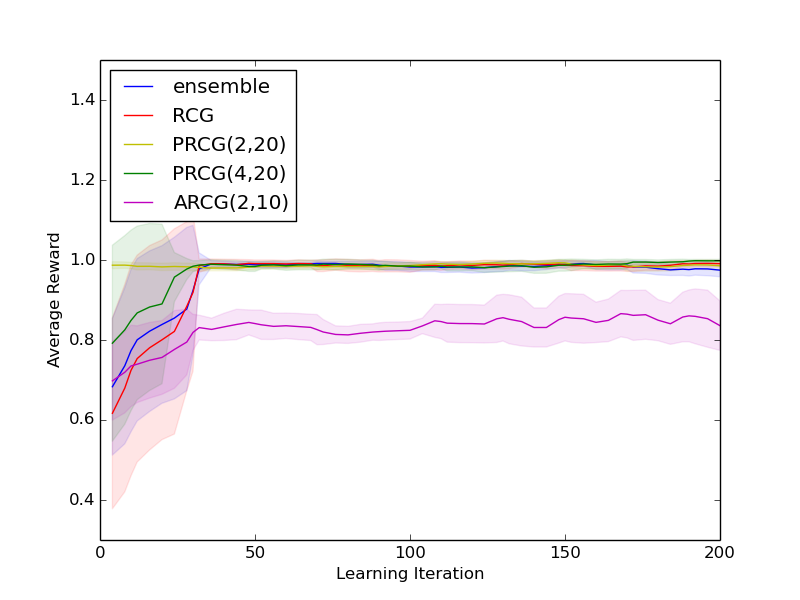}}%
    \subfigure[Experiment 2 (Mid, F)]{
        \label{fig:door_exp2}
        \includegraphics[width=0.3\linewidth]{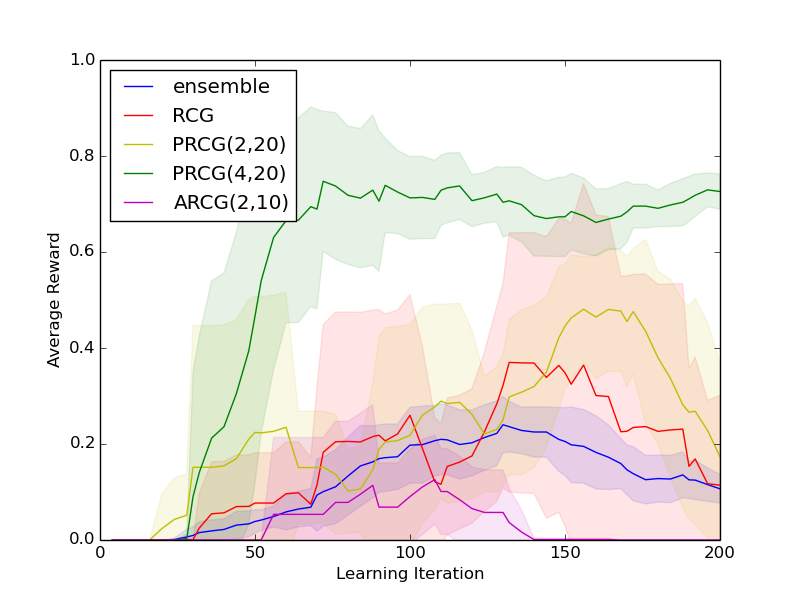}}
    \subfigure[Experiment 3 (Far, F)]{
        \label{fig:door_exp3}
        \includegraphics[width=0.3\linewidth]{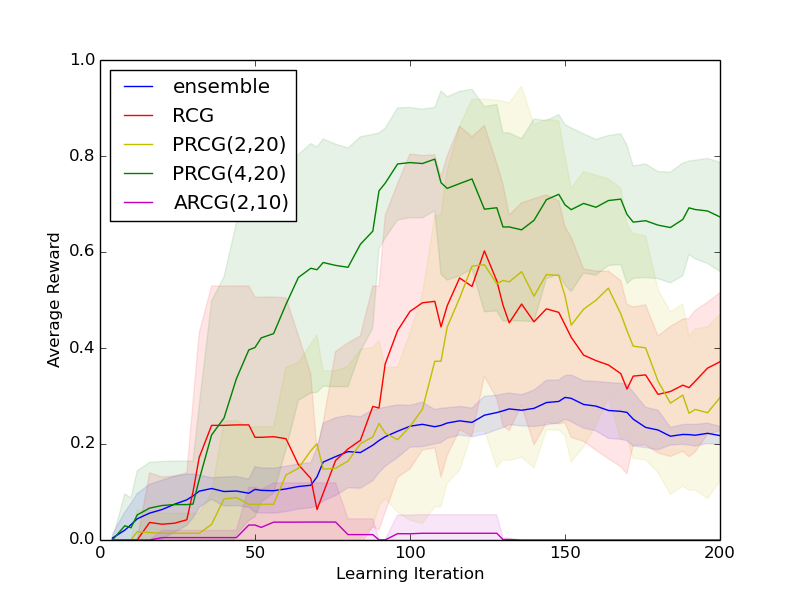}}
    \subfigure[Experiment 4 (Near, V)]{
        \label{fig:door_exp4}
        \includegraphics[width=0.3\linewidth]{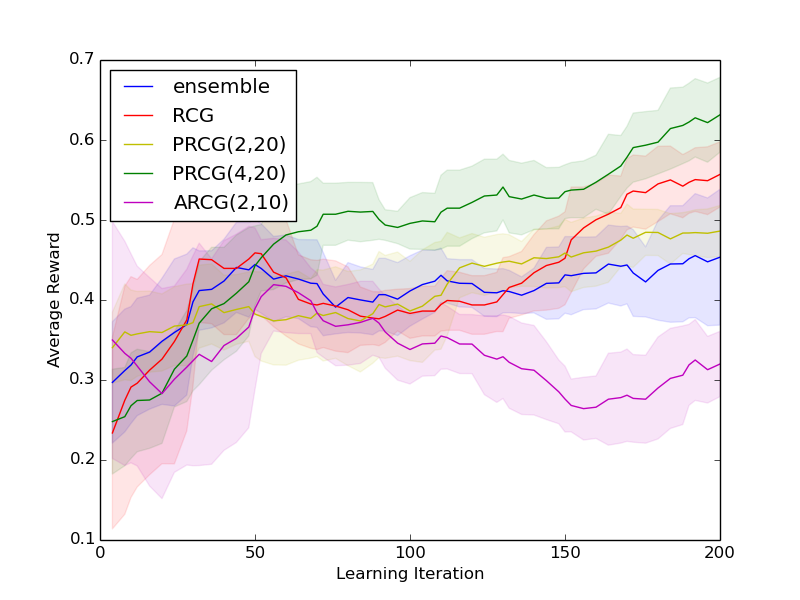}}%
    \subfigure[Experiment 5 (Mid, V)]{
        \label{fig:door_exp5}
        \includegraphics[width=0.3\linewidth]{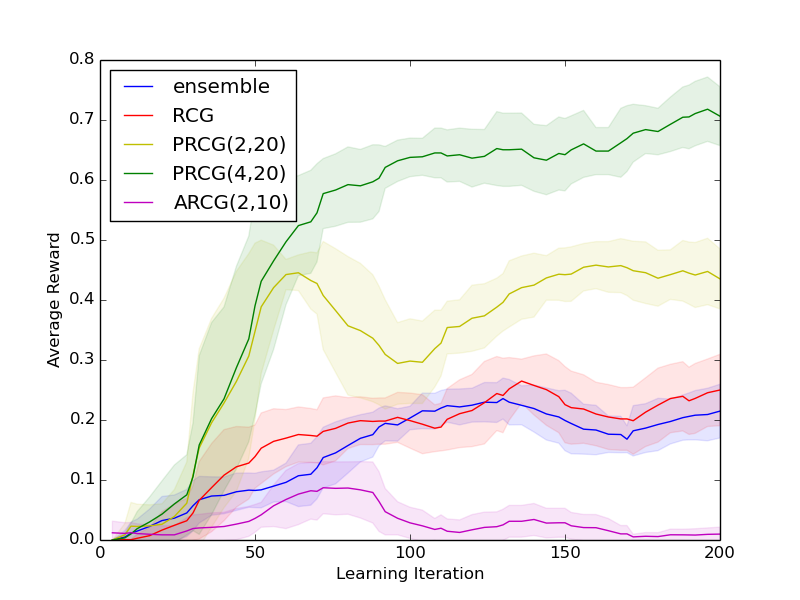}}
    \subfigure[Experiment 6 (Far, V)]{
        \label{fig:door_exp6}
        \includegraphics[width=0.3\linewidth]{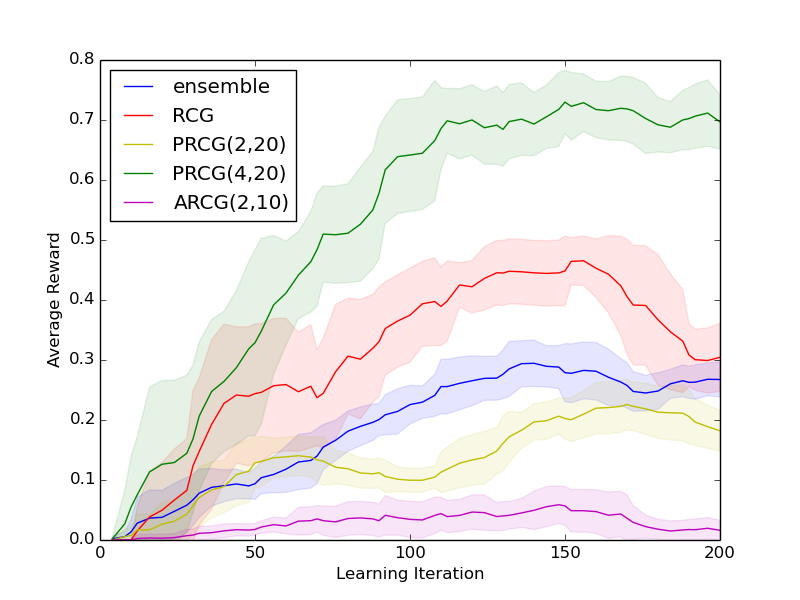}}
    \captionof{figure}{Learning curve for the door opening task}
    \vspace{-15pt}
    \label{fig:door_lc}
\end{figure*}

\begin{figure*}[t!]
    \centering
    \subfigure[Experiment 1 (Near, F)]{
        \label{fig:key_exp1}
        \includegraphics[width=0.3\linewidth]{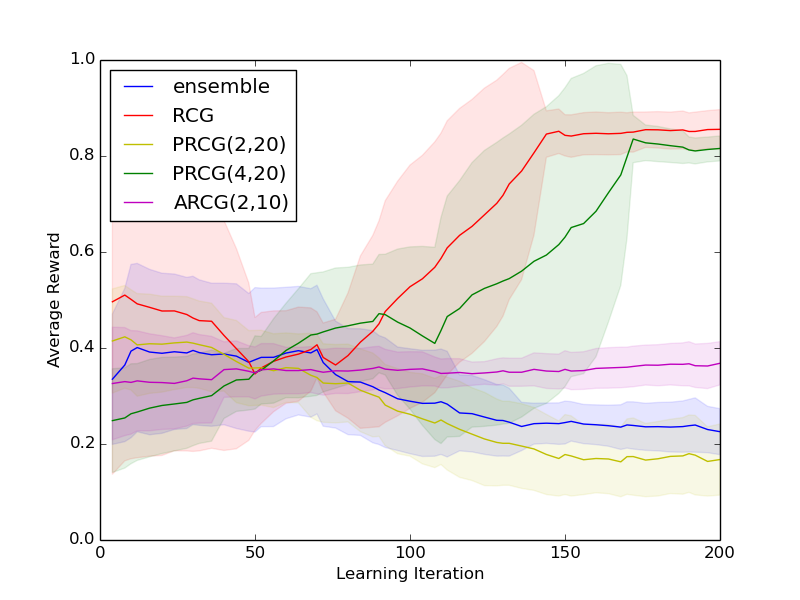}}%
    \subfigure[Experiment 2 (Mid, F)]{
        \label{fig:key_exp2}
        \includegraphics[width=0.3\linewidth]{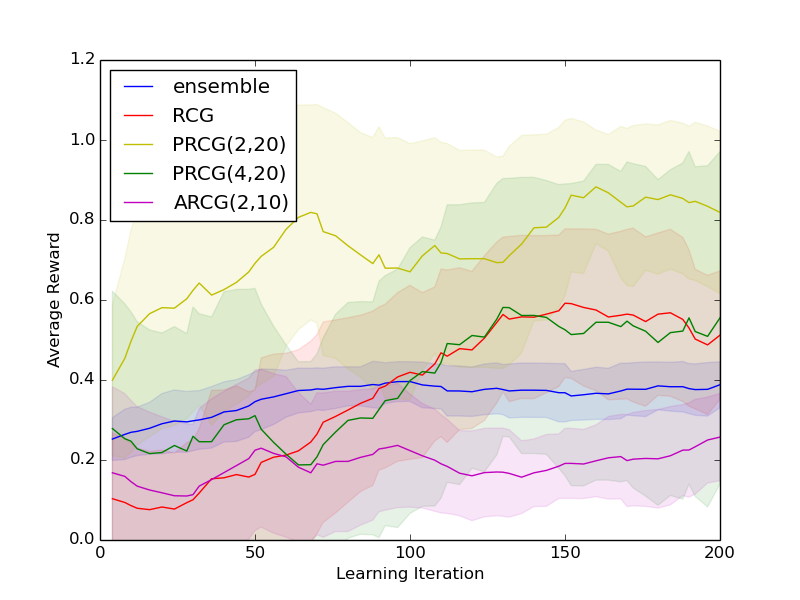}}
    \subfigure[Experiment 3 (Far, F)]{
        \label{fig:key_exp3}
        \includegraphics[width=0.3\linewidth]{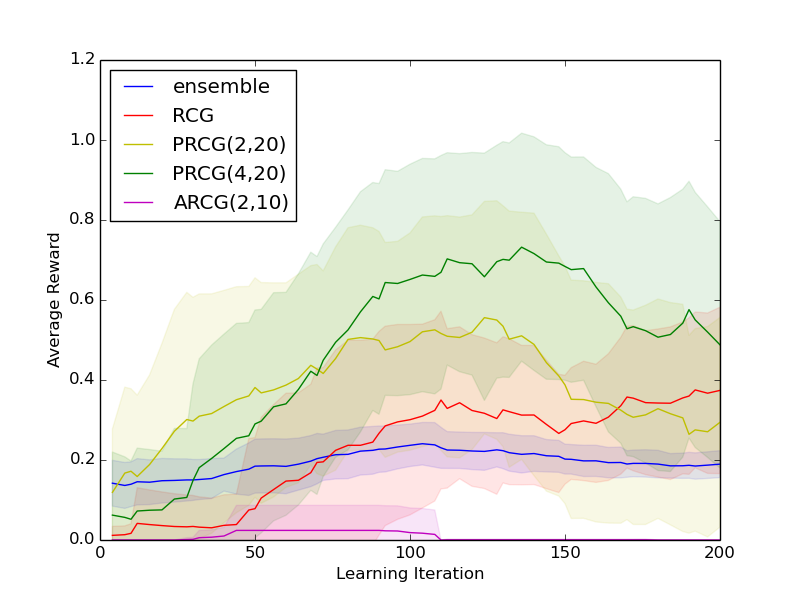}}
    \subfigure[Experiment 4 (Near, V)]{
        \label{fig:key_exp4}
        \includegraphics[width=0.3\linewidth]{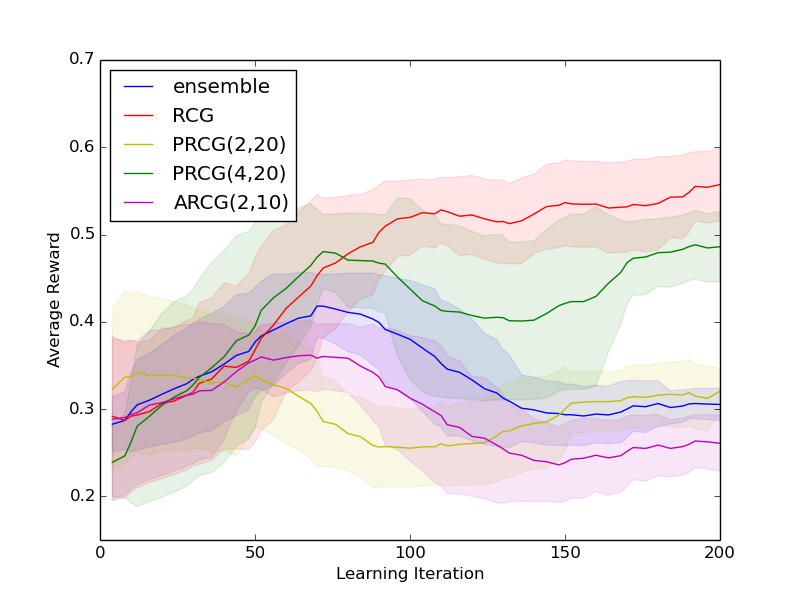}}%
    \subfigure[Experiment 5 (Mid, V)]{
        \label{fig:key_exp5}
        \includegraphics[width=0.3\linewidth]{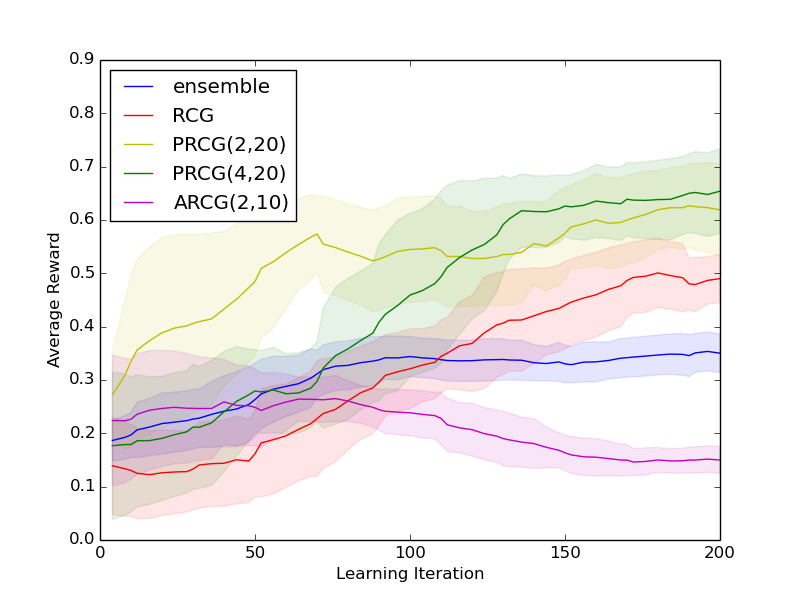}}
    \subfigure[Experiment 6 (Far, V)]{
        \label{fig:key_exp6}
        \includegraphics[width=0.3\linewidth]{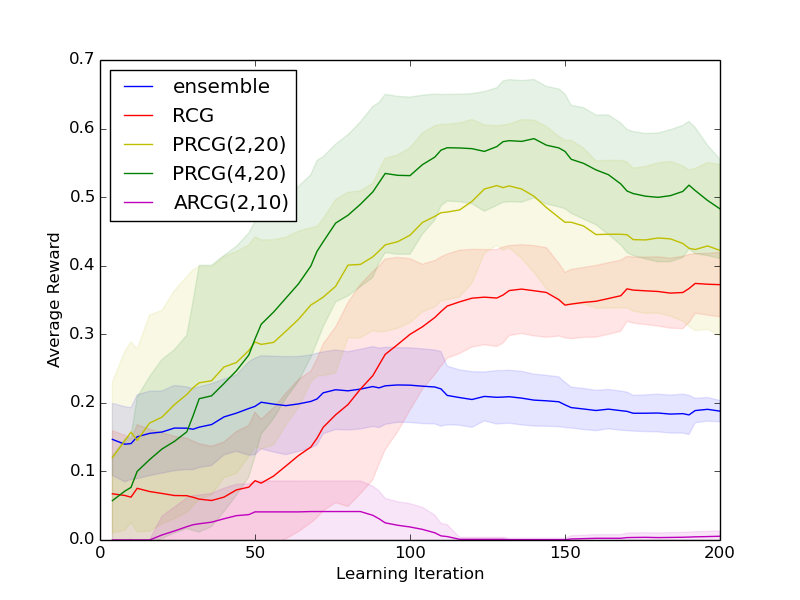}}
    \captionof{figure}{Learning curve for the key insertion task}
    \vspace{-15pt}
    \label{fig:key_lc}
\end{figure*}

Fig. \ref{fig:grasp_lc} to Fig. \ref{fig:key_lc} show the learning curves for the three tasks.
Fig. \ref{fig:grasp_exp1} to Fig. \ref{fig:grasp_exp3}, Fig. \ref{fig:door_exp1} to Fig. \ref{fig:door_exp3}, and Fig. \ref{fig:key_exp1} to Fig. \ref{fig:key_exp3} show the learning curves for Experiment 1 to 3. 
For Experiment 1, the performance of  RCG, PRCG(2,20), and PRCG(4,20) do not show much difference in the grasping and door opening tasks. This is reasonable since all the models learn from the initial states closer to the target at the beginning, the sampling nearby process are more likely to cover the states under this experiment.
In the key insertion task, RCG and PRCG(4,20) also perform similarly.
For Experiment 2 and 3, the improvement of PRCG(4,20) becomes apparent in most cases. Since each actor trained by PRCG needs to find a more general policies to adapt the exchanged critic, the policy is less likely to overfit to the biased initial state distribution.
For Experiment 1 to 3, PRCG(2,20) also improves in some cases but mostly does not show much difference from RCG.

Fig. \ref{fig:grasp_exp4} to Fig. \ref{fig:grasp_exp6}, Fig. \ref{fig:door_exp4} to Fig. \ref{fig:door_exp6}, and Fig. \ref{fig:key_exp4} to Fig. \ref{fig:key_exp6} show the learning curves for Experiment 4 to 6. 
Since the initial pose of the robot arm is variable, we can now observe generality of the models.
Experiment 4 to 6 show that the generality of PRCG(2,20) and PRCG(4,20) is better than RCG in the grasping task, even when the initial state is already near to the goal state. 
In the door opening and key insertion tasks, PRCG(2,20) and PRCG(4,20) outperforms RCG in most cases.
Obviously, by exchange of more critics, PRCG(4,20) can achieve a higher performance in a smaller number of training iterations. 
These results can demonstrate that the parallelized technique with a proper choice of the number of models can not only increase the expansion rate of the learned initial positions of the gripper but also prevent the policies from only learning from some specific region.
However, PRCG(2,20) sometimes does not outperform or even becomes worse than RCG.
The undesired performance of PRCG(2,20) in Experiment 1 to 6 might be due to the following 2 reasons, and they are both more likely to happen when the model number $m$ is small: 
\begin{itemize}[leftmargin=*]
    \item The 2 models cover a similar range of initial states, so the model does not explore much. 
    \item Although the swap frequency $K=20$ is beneficial for learning most of the time, a model might not have learned to some extent after 20 iterations, and swapping will instead make learning unstable.
\end{itemize}

It is worth notice that the effect of different number of models trained by PRCG is more obvious when the gripper is further from the target object. Learning curves of Experiment 2, 3, 5 and 6 in Fig. \ref{fig:grasp_lc} to Fig. \ref{fig:key_lc} show that PRCG(4,20) outperforms RCG and PRCG(2,20) in most cases. 
When training with more models simultaneously, each actor can find a more general policy, the initial states near the target can be learned in a smaller number of training iterations.
Therefore, initial states further from the target can be included in the good start pool faster, and the range of the learned initial positions will expand more. 
Note that here we only train our models for a fix number of iterations, and the performance might be increased as the training continues. However, we can still observe the improvement of our methods under this condition.

We also compare our approach with an ensemble and an asynchronous versions. 
For the ensemble version, we train 2 RCG models simultaneously without swapping and combine the policies of these 2 models.
The asynchronous version is implemented just like A3C~\citep{mnih2016asynchronous}, and we call it asynchronous reverse curriculum generation (ARCG). 
In our experiments, we train ARCG with two models and update each model every 10 iterations (ARCG(2, 10)). 
The ensemble experiments fail in most cases since some RCG models are not well learned and decrease the performance of the combined models.
For the asynchronous experiments, even using the best hyperparameters we found so far, ARCG still performs poorly in all tasks as the purple curves shown in Fig. \ref{fig:grasp_lc} to Fig. \ref{fig:key_lc}.
We argue that this happens because in ARCG, the global policy needs to periodically send its parameters to each local policy, while PRCG can let each policy develop their own strategies that might have very different network parameters among them.

\subsubsection{The effects of different training strategies for PRCG}

Besides swapping critics, there are other training strategies that can produce a synergistic effect for PRCG.
In our experiments, we compare the following four strategies: 
\begin{itemize}[leftmargin=*]
    \item Train $m=4$ RCG models simultaneously with separate initial state pools and swap critics at $K=20$ (PRCG(4,20)).
    \item Train $m=4$ RCG models simultaneously with separate initial state pools and swap initial state pools at $K=20$ (PRCG(4,20)-i). 
    \item Train $m=4$ RCG models simultaneously with a common initial state pool and swap critics at $K=20$ (PRCG(4,20)-ci). 
    \item Train $m=4$ RCG models simultaneously with a common initial state pool and do not swap critics (RCG(4)-ci).
\end{itemize}

\begin{figure*}[t!]
    \centering
    \subfigure[PRCG(4,20)]{
        \label{fig:grasp_PRCG_4_20_gp}
        \includegraphics[width=0.23\linewidth]{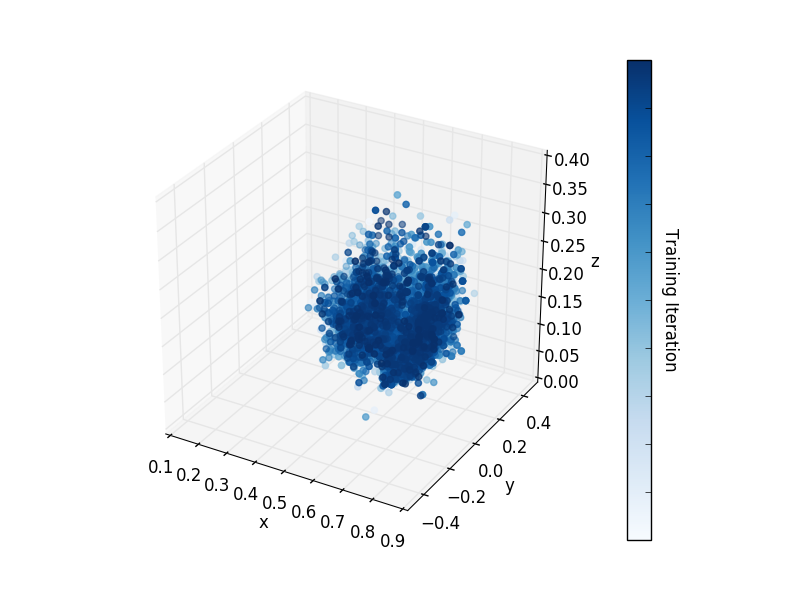}
    }
    \subfigure[PRCG(4,20)-i]{
        \label{fig:grasp_PRCG_4_20_i_gp}
        \includegraphics[width=0.23\linewidth]{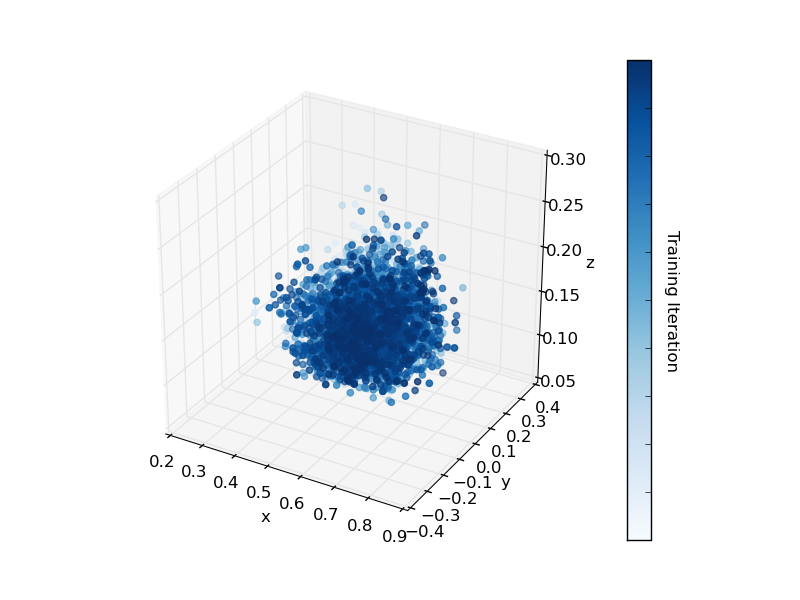}}
    \subfigure[PRCG(4,20)-ci]{
        \label{fig:grasp_PRCG_4_20_ci_gp}
        \includegraphics[width=0.23\linewidth]{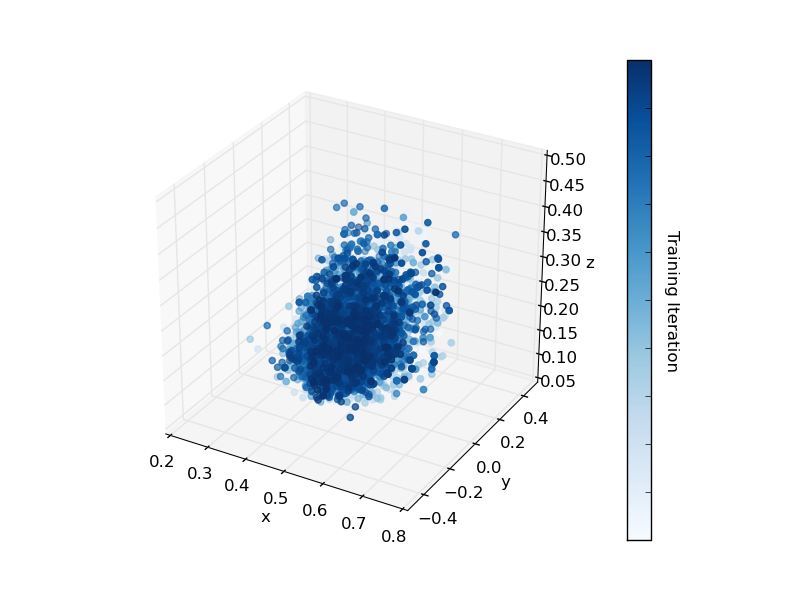}}
    \subfigure[RCG(4)-ci]{
        \label{fig:grasp_RCG_4_ci_gp}
        \includegraphics[width=0.23\linewidth]{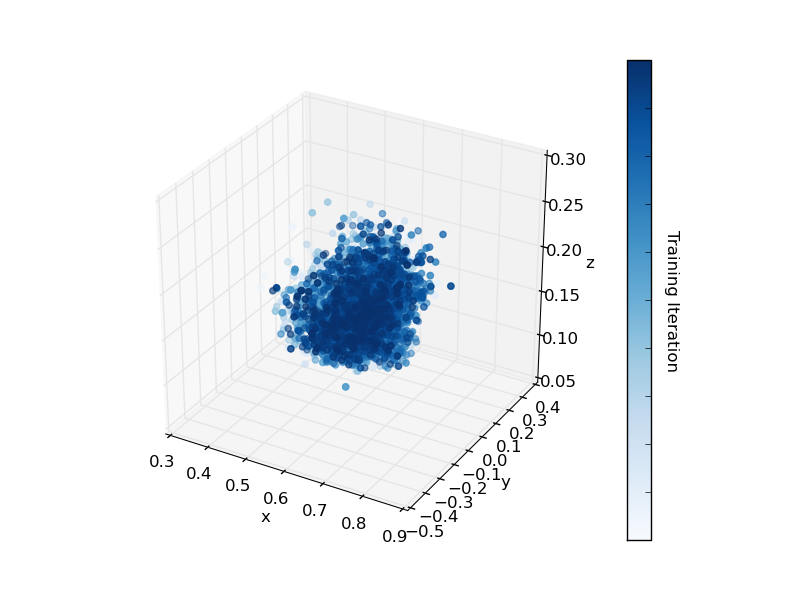}}
    \captionof{figure}{Learned initial positions of different PRCG training strategies (Grasping task; The target object is approximately at (0.5, 0, -0.1).)}
    \vspace{-15pt}
    \label{fig:grasp_gp_2}
\end{figure*}

\begin{figure*}[t!]
    \centering
    \subfigure[PRCG(4,20)]{
        \label{fig:door_PRCG_4_20_gp}
        \includegraphics[width=0.23\linewidth]{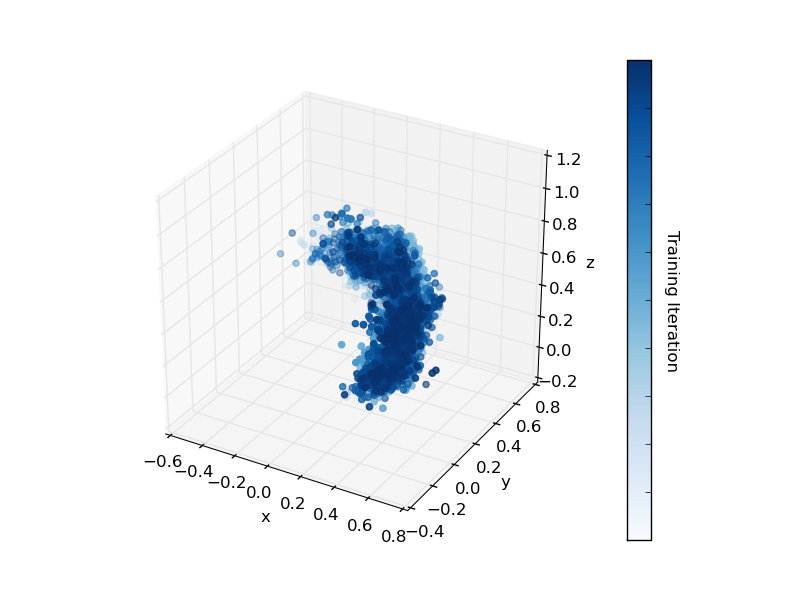}
    }
    \subfigure[PRCG(4,20)-i]{
        \label{fig:door_PRCG_4_20_i_gp}
        \includegraphics[width=0.23\linewidth]{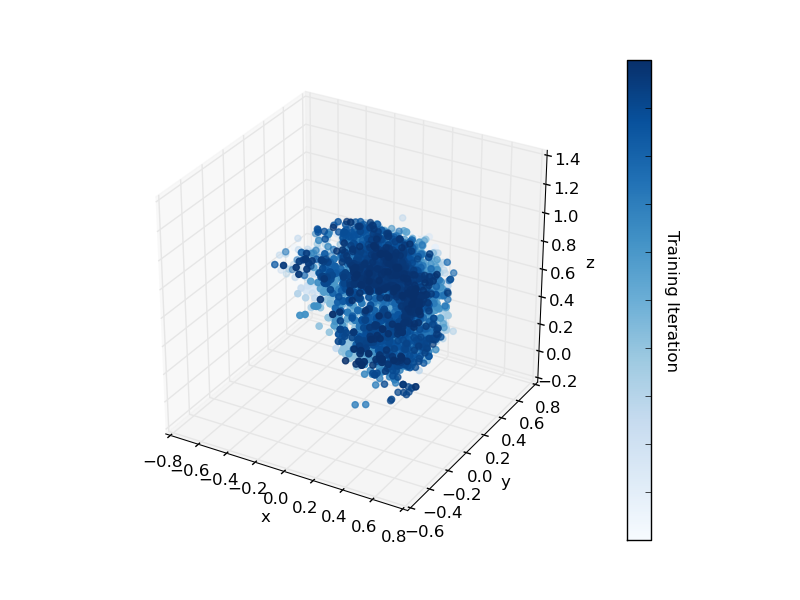}}
    \subfigure[PRCG(4,20)-ci]{
        \label{fig:door_PRCG_4_20_ci_gp}
        \includegraphics[width=0.23\linewidth]{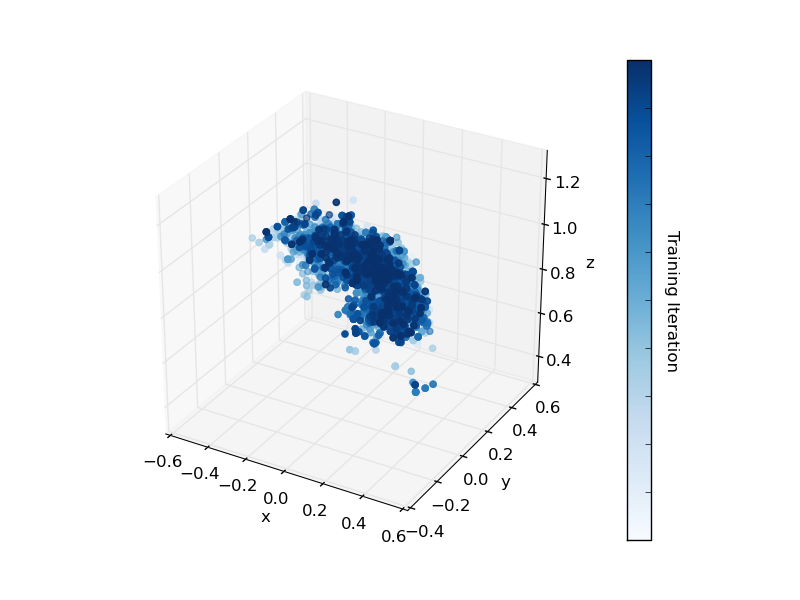}}
    \subfigure[RCG(4)-ci]{
        \label{fig:door_RCG_4_ci_gp}
        \includegraphics[width=0.23\linewidth]{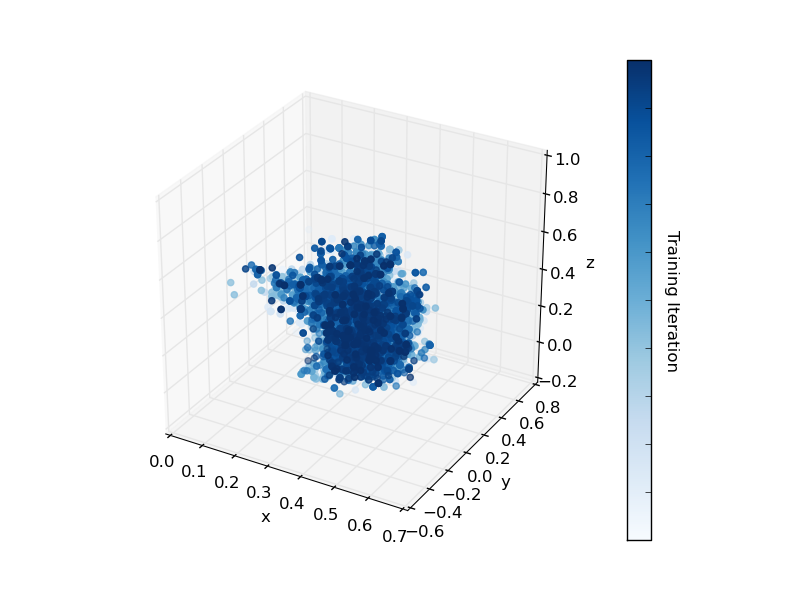}}
    \captionof{figure}{Learned initial positions of different PRCG training strategies (Door opening task; The target object is approximately at (0.35, 0.025, 0.2).)}
    \vspace{-15pt}
    \label{fig:door_gp_2}
\end{figure*}

\begin{figure*}[t!]
    \centering
    \subfigure[PRCG(4,20)]{
        \label{fig:key_PRCG_4_20_gp}
        \includegraphics[width=0.23\linewidth]{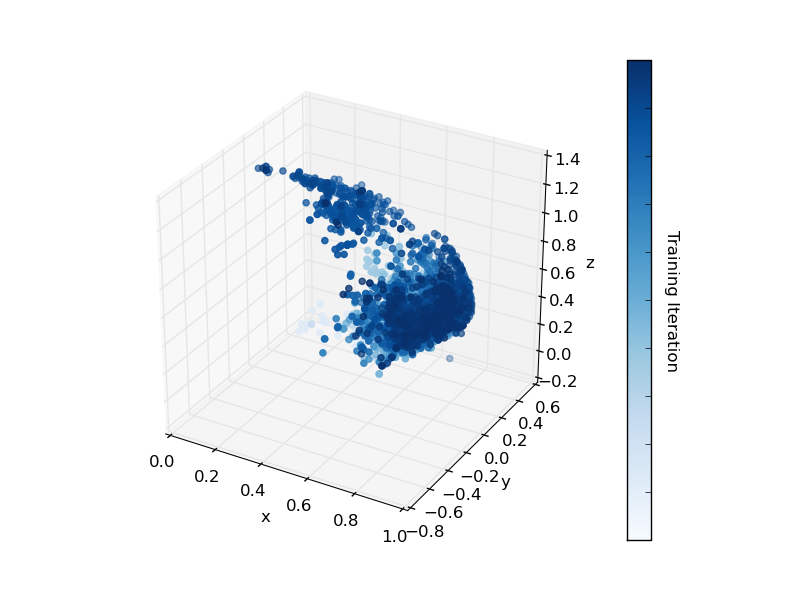}
    }
    \subfigure[PRCG(4,20)-i]{
        \label{fig:key_PRCG_4_20_i_gp}
        \includegraphics[width=0.23\linewidth]{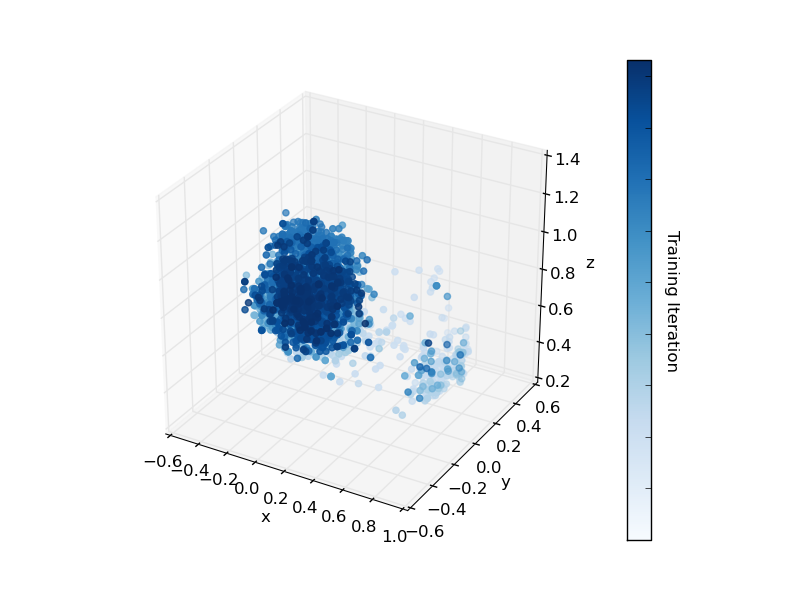}}
    \subfigure[PRCG(4,20)-ci]{
        \label{fig:key_PRCG_4_20_ci_gp}
        \includegraphics[width=0.23\linewidth]{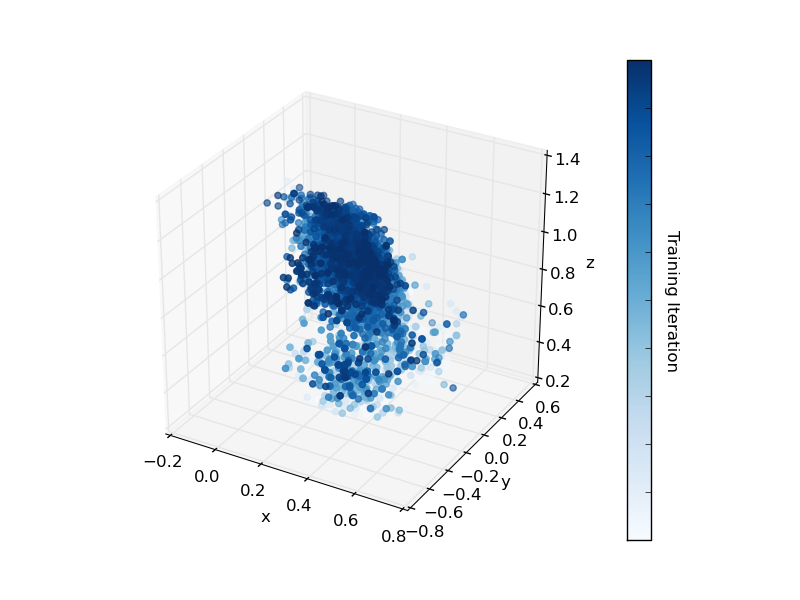}}
    \subfigure[RCG(4)-ci]{
        \label{fig:key_RCG_4_ci_gp}
        \includegraphics[width=0.23\linewidth]{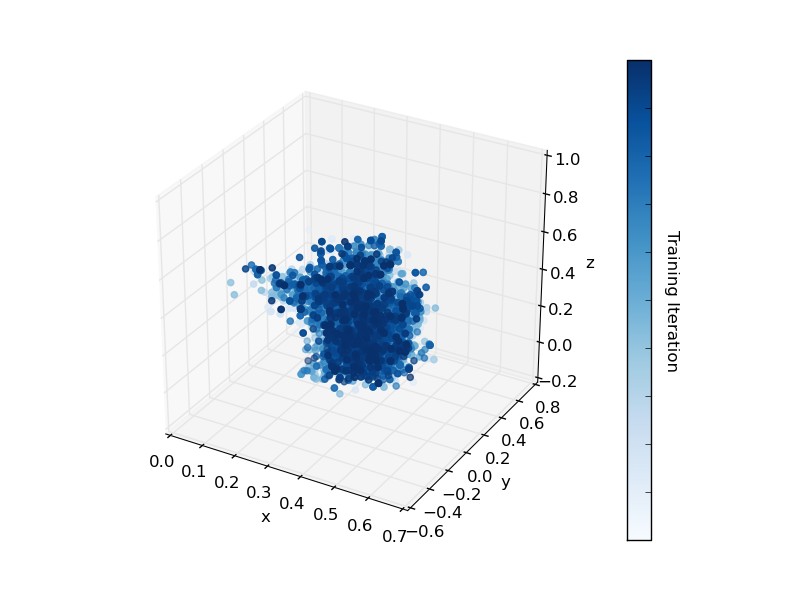}}
    \captionof{figure}{Learned initial positions of different PRCG training strategies (Key insertion task; The target object is approximately at (0.75, 0, 0.5).)}
    \vspace{-15pt}
    \label{fig:key_gp_2}
\end{figure*}

Fig. \ref{fig:grasp_gp_2} to Fig. \ref{fig:key_gp_2} show the learned initial positions of the gripper collected by the models trained with PRCG(4,20), PRCG(4,20)-i, PRCG(4,20)-ci, and RCG(4)-ci. 
In all tasks, RCG(4)-ci does not explore as much as other models do. 
In the grasping and key insertion tasks, both PRCG(4,20) and PRCG(4,20)-ci explore more than PRCG(4,20)-i does. 
In the door opening task, both PRCG(4,20)-ci and PRCG(4,20)-i explore more than PRCG(4,20). 
These results align with the analysis mentioned in Section \ref{subsec:diff_prcg}. 
Since swapping critics leads to faster exploration, the learned initial states of PRCG(4,20) can expand faster in most cases.
Additionally, among all models, PRCG(4,20) is the one that concentrates most on the desired region. 
This is especially obvious in the key insertion task. 
Therefore, PRCG(4,20) is most likely to succeed from the desired initial states.

\begin{figure*}[t!]
    \centering
    \subfigure[Experiment 1 (Near, F)]{
        \label{fig:grasp_exp1_2}
        \includegraphics[width=0.3\linewidth]{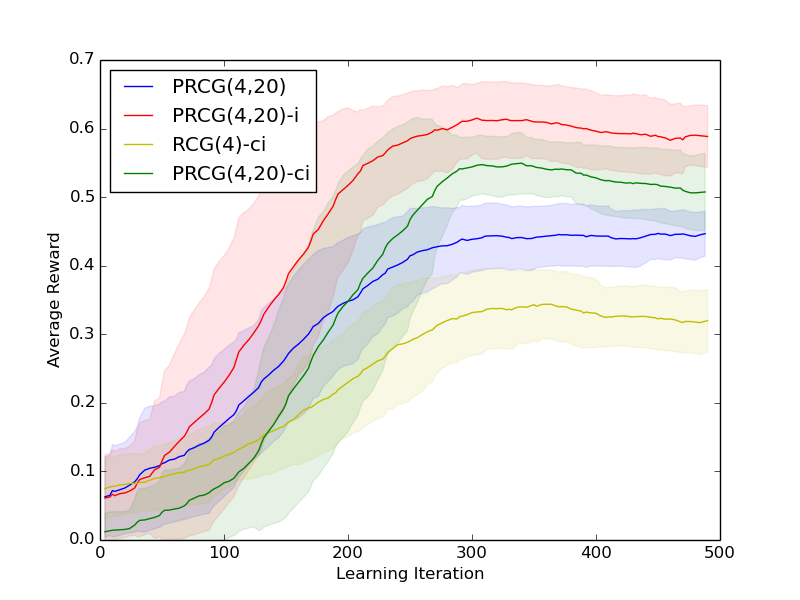}}%
    \subfigure[Experiment 2 (Mid, F)]{
        \label{fig:grasp_exp2_2}
        \includegraphics[width=0.3\linewidth]{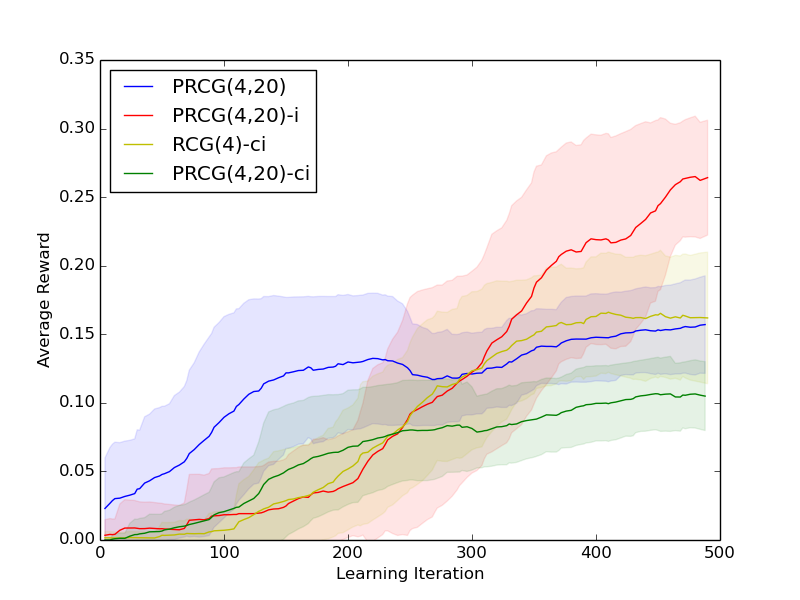}}
    \subfigure[Experiment 3 (Far, F)]{
        \label{fig:grasp_exp3_2}
        \includegraphics[width=0.3\linewidth]{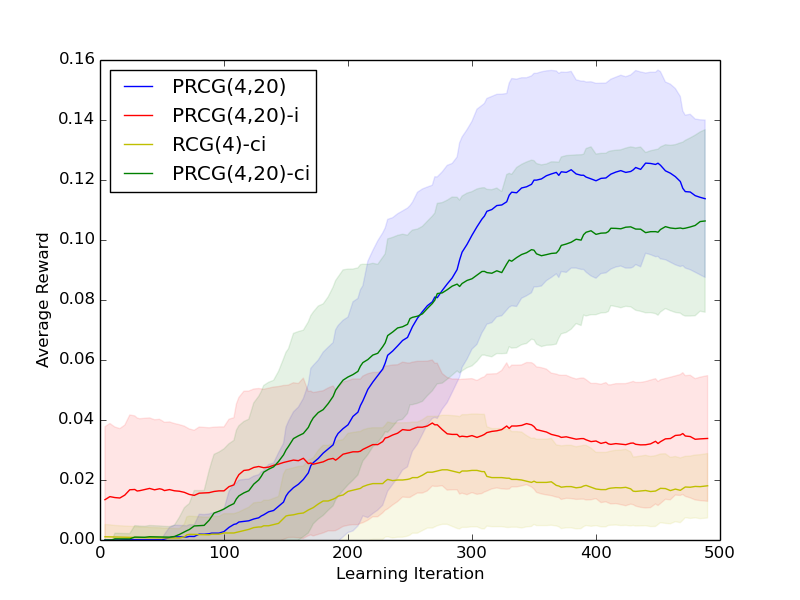}}
    \subfigure[Experiment 4 (Near, V)]{
        \label{fig:grasp_exp4_2}
        \includegraphics[width=0.3\linewidth]{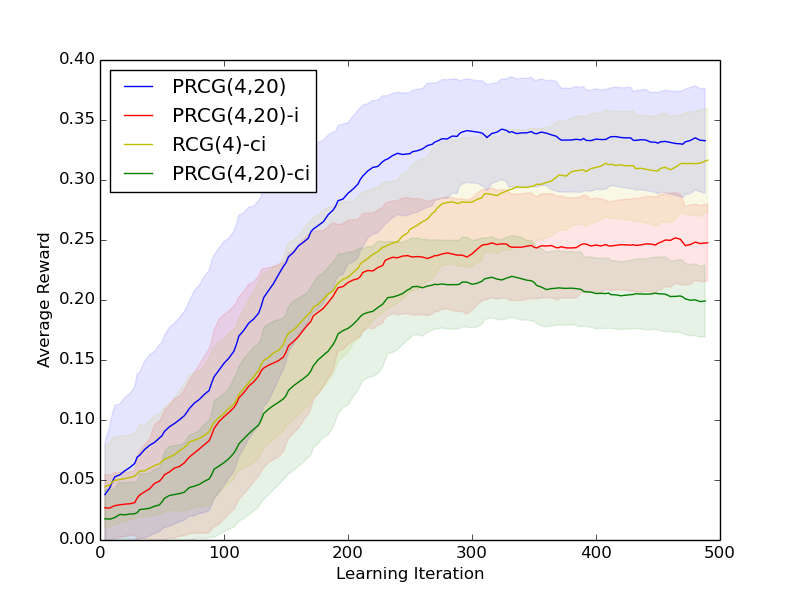}}%
    \subfigure[Experiment 5 (Mid, V)]{
        \label{fig:grasp_exp5_2}
        \includegraphics[width=0.3\linewidth]{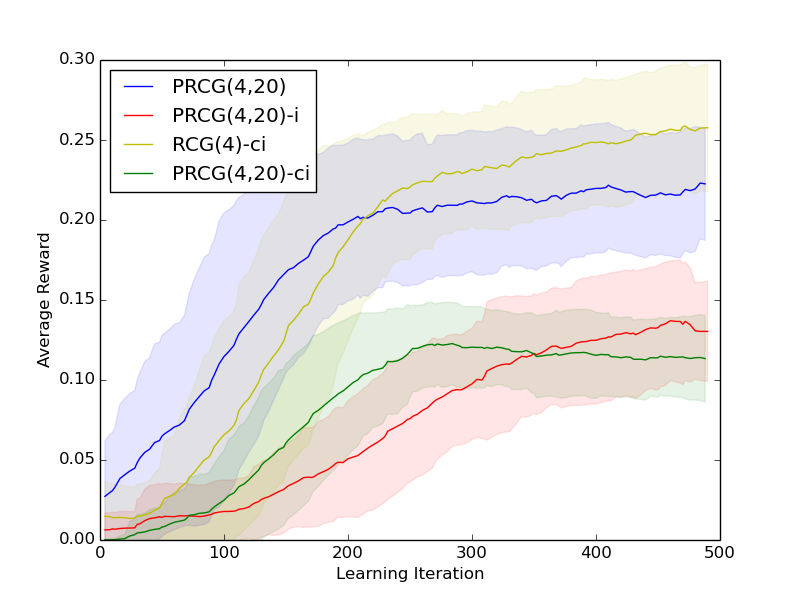}}
    \subfigure[Experiment 6 (Far, V)]{
        \label{fig:grasp_exp6_2}
        \includegraphics[width=0.3\linewidth]{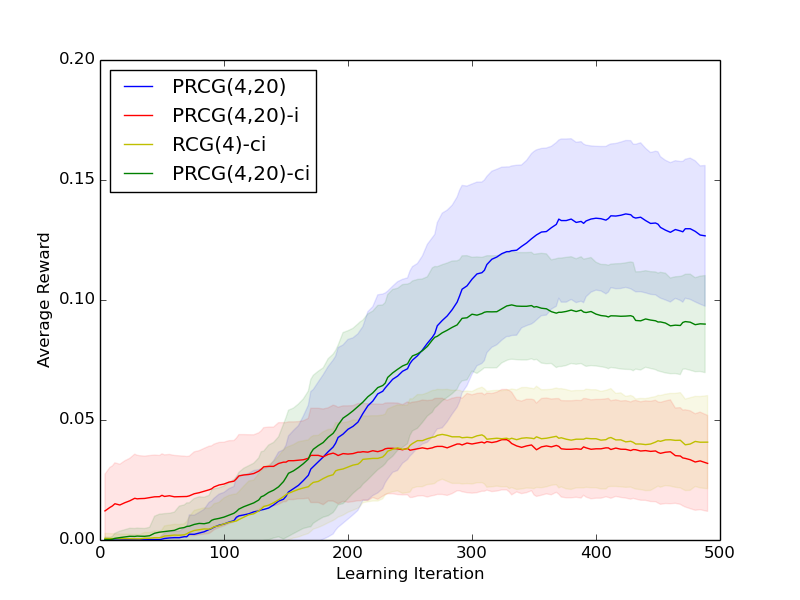}}
    \captionof{figure}{Comparison of different PRCG training strategies for the grasping task}
    \vspace{-15pt}
    \label{fig:grasp_lc_2}
\end{figure*}

\begin{figure*}[t!]
    \centering
    \subfigure[Experiment 1 (Near, F)]{
        \label{fig:door_exp1_2}
        \includegraphics[width=0.3\linewidth]{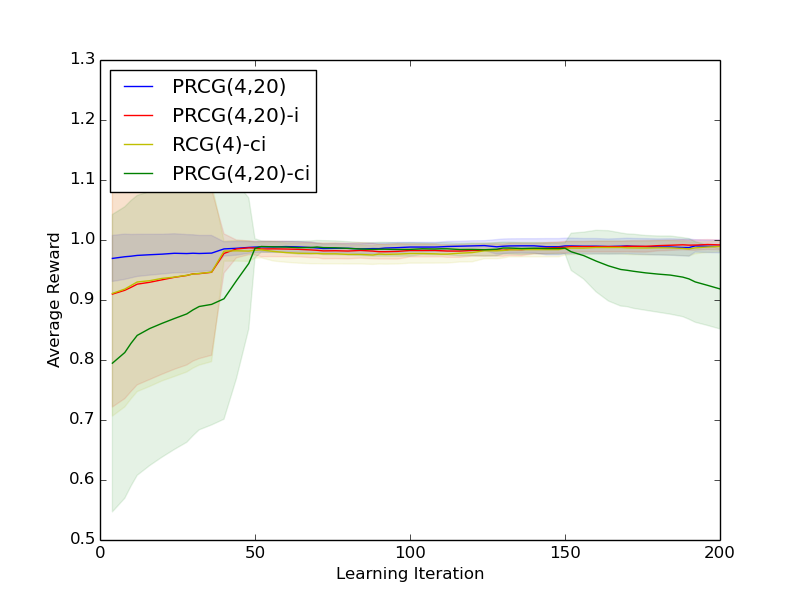}}%
    \subfigure[Experiment 2 (Mid, F)]{
        \label{fig:door_exp2_2}
        \includegraphics[width=0.3\linewidth]{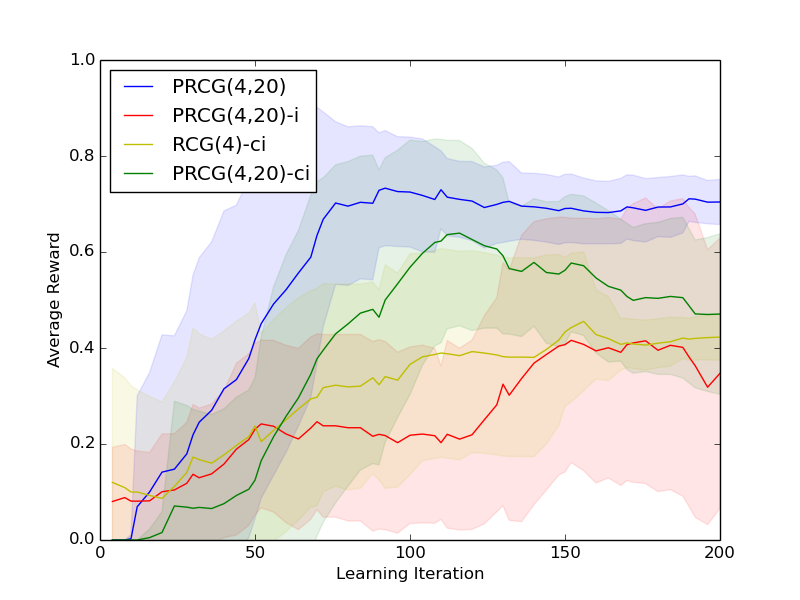}}
    \subfigure[Experiment 3 (Far, F)]{
        \label{fig:door_exp3_2}
        \includegraphics[width=0.3\linewidth]{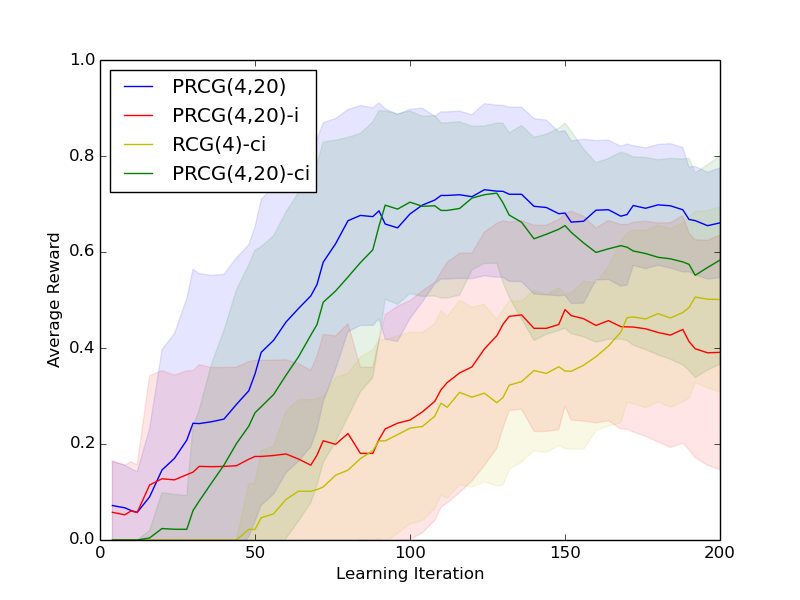}}
    \subfigure[Experiment 4 (Near, V)]{
        \label{fig:door_exp4_2}
        \includegraphics[width=0.3\linewidth]{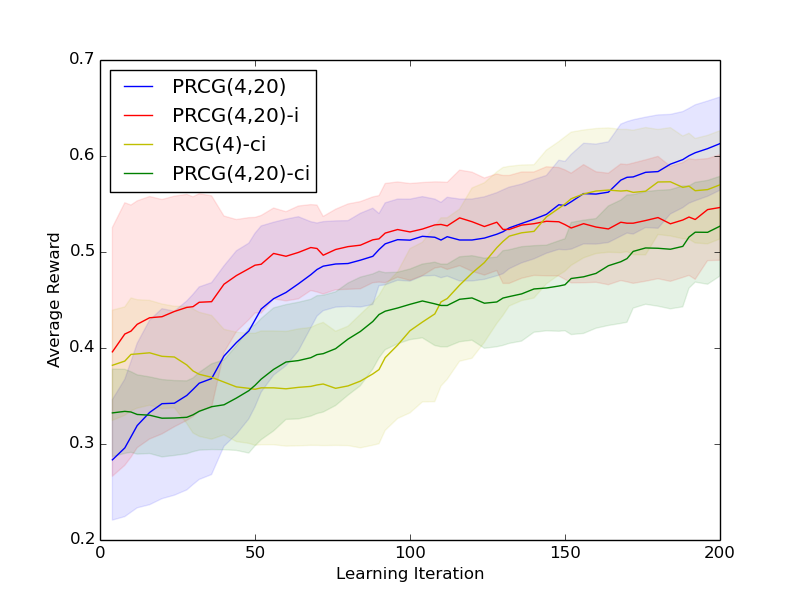}}%
    \subfigure[Experiment 5 (Mid, V)]{
        \label{fig:door_exp5_2}
        \includegraphics[width=0.3\linewidth]{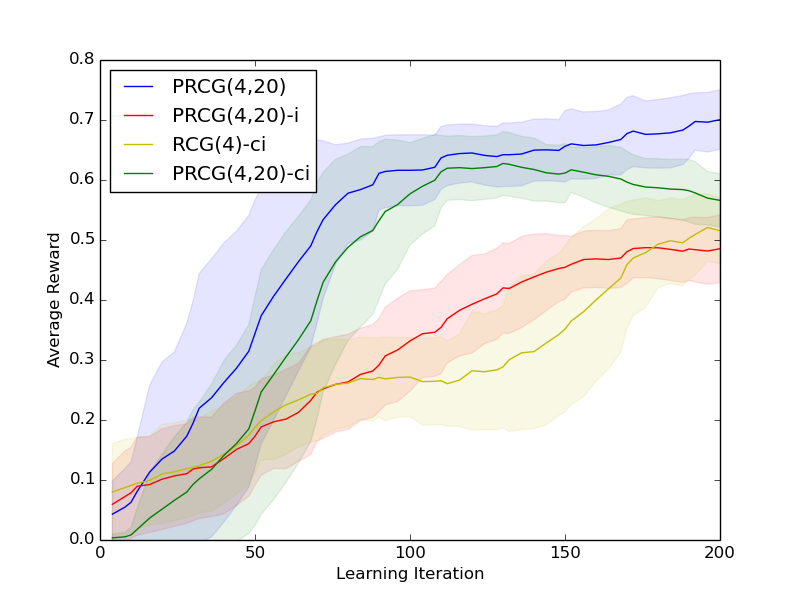}}
    \subfigure[Experiment 6 (Far, V)]{
        \label{fig:door_exp6_2}
        \includegraphics[width=0.3\linewidth]{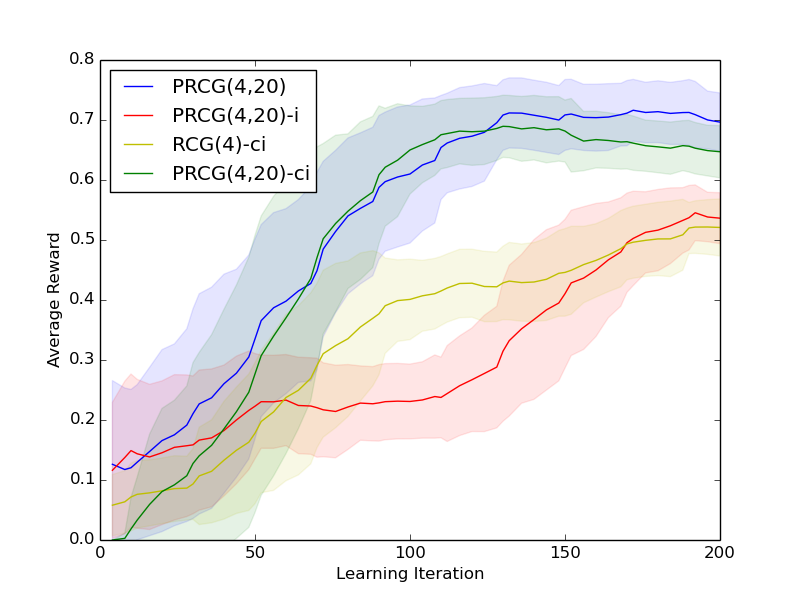}}
    \captionof{figure}{Comparison of different PRCG training strategies for the door opening task}
    \vspace{-15pt}
    \label{fig:door_lc_2}
\end{figure*}

\begin{figure*}[t!]
    \centering
    \subfigure[Experiment 1 (Near, F)]{
        \label{fig:key_exp1_2}
        \includegraphics[width=0.3\linewidth]{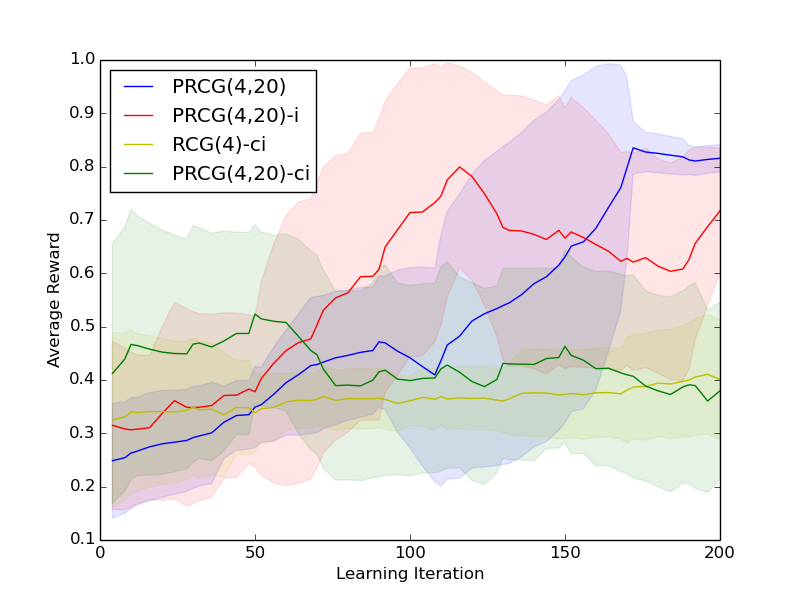}}%
    \subfigure[Experiment 2 (Mid, F)]{
        \label{fig:key_exp2_2}
        \includegraphics[width=0.3\linewidth]{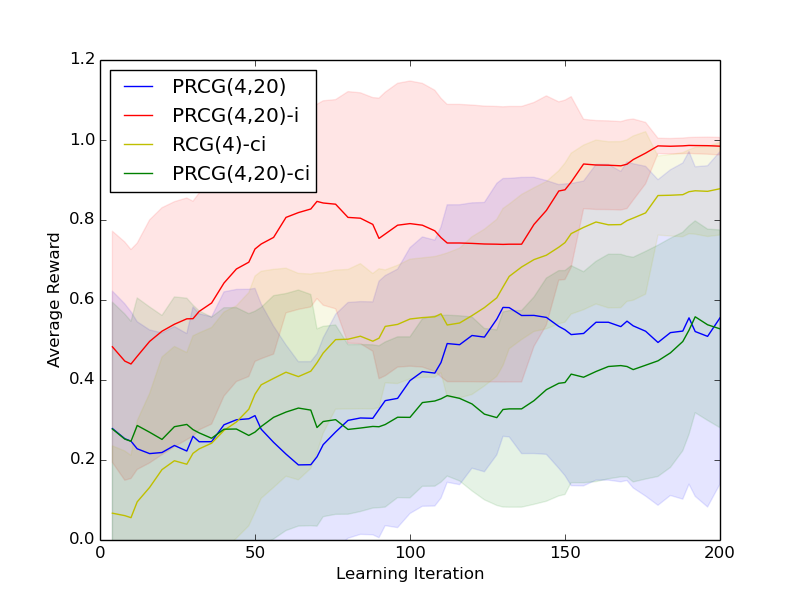}}
    \subfigure[Experiment 3 (Far, F)]{
        \label{fig:key_exp3_2}
        \includegraphics[width=0.3\linewidth]{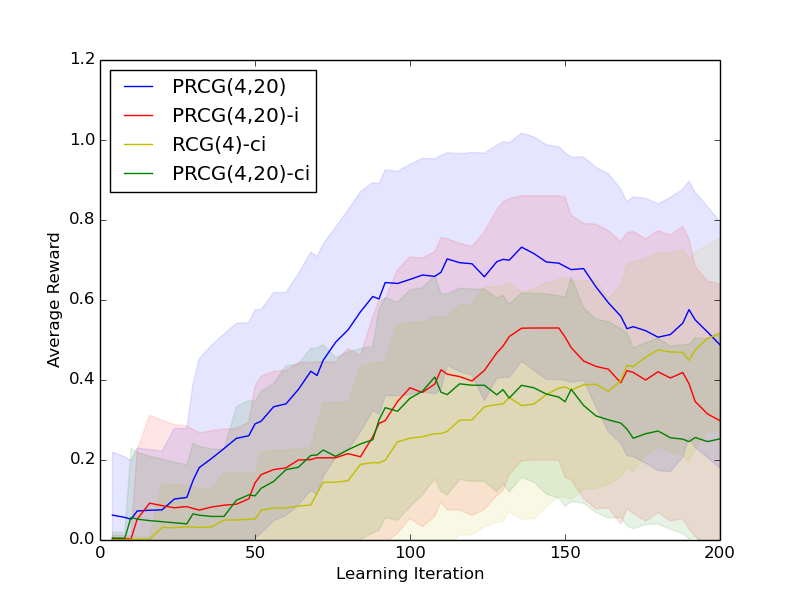}}
    \subfigure[Experiment 4 (Near, V)]{
        \label{fig:key_exp4_2}
        \includegraphics[width=0.3\linewidth]{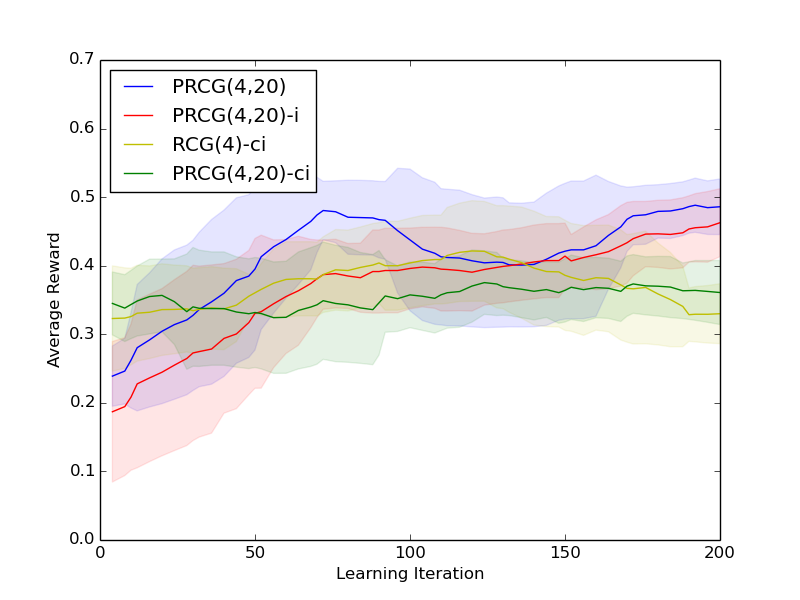}}%
    \subfigure[Experiment 5 (Mid, V)]{
        \label{fig:key_exp5_2}
        \includegraphics[width=0.3\linewidth]{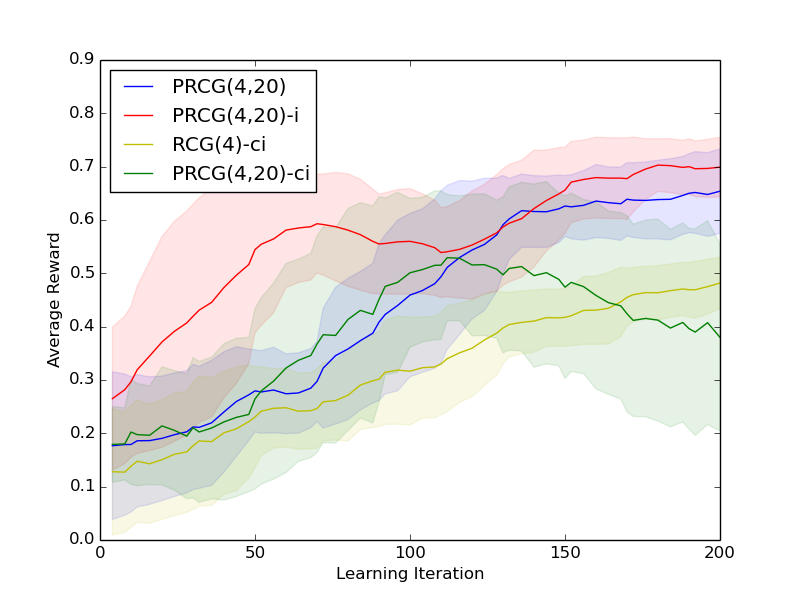}}
    \subfigure[Experiment 6 (Far, V)]{
        \label{fig:key_exp6_2}
        \includegraphics[width=0.3\linewidth]{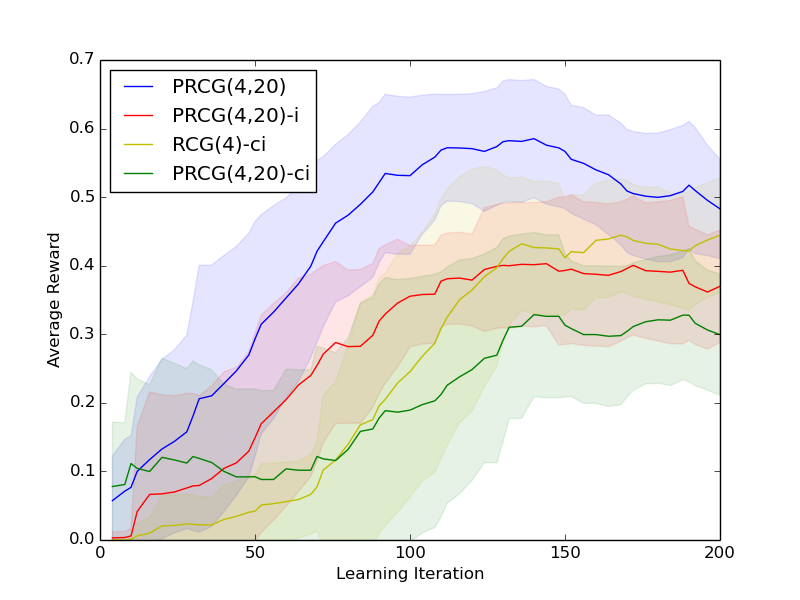}}
    \captionof{figure}{Comparison of different PRCG training strategies for the key insertion task}
    \vspace{-15pt}
    \label{fig:key_lc_2}
\end{figure*}

Fig. \ref{fig:grasp_lc_2} to Fig. \ref{fig:key_lc_2} show the learning curves of different training strategies for the three tasks. 
In all tasks, it can be observed that PRCG(4,20) outperforms PRCG(4,20)-i for Experiment 3 and 6, which are the most difficult cases. 
For Experiment 1, 2, 4, and 5, PRCG(4,20)-i sometimes performs better than PRCG(4,20), but overall PRCG(4,20) can get a higher reward. 
These results show that swapping critics can learn from the harder initial states faster and still maintain good performance from the easier initial states.

In general, PRCG(4,20)-ci and RCG(4)-ci do not outperform PRCG(4,20). 
Additionally, they are sometimes unstable during training since their performance highly depends on the quality of the initial state pool. 
If the good starts collected by each critic are in a very similar range, then all models will be trained from similar initial states and have seldom synergistic effect. 
This is more likely to happen when all models share a common initial state pool. 
If the good starts collected by each critic are separate but with some intersection, then each model can benefit from the initial states collected by others to increase exploration. 
If the good starts collected by each critic do not have enough intersection, then a model might be trained from initial states that are too difficult for it. 
This will make training unstable.

\subsubsection{Effects of the parallelized approach in PRCG on other AC based RL algorithms}

We also test our method on another AC based RL algorithm which might benefit from the parallelized approach in PRCG. 
An important characteristic of RCG is the adapted initial state distribution, which makes the policy observe strong learning signals more often.
Therefore, we choose RL methods with adapted initial state distribution since the agent is more likely to receive useful learning signals.
Since few methods automatically generate adapted initial state distribution like RCG does, we develop a simple method that works only for simple environments.
This method is to concatenate a random noise vector to the input vector as the adapted initial state distribution, and this noise vector can also be used for exploration.
We choose to apply this method to DDPG~\citep{lillicrap2015continuous} without adding noise to the actor output, and we call this new method DDPG-AI.
All the hyperparameters used are the same as the ones set in OpenAI Baselines~\citep{baselines}.
The algorithm is applied to \textit{InvertedPendulum} and \textit{InvertedDoublePendulum}, which are two non-sparse reward environments in OpenAI Gym~\citep{1606.01540}.
Here we compare three variations of DDPG for each environment: DDPG, DDPG with adapted initial state distribution (DDPG-AI), parallelized DDPG with adapted initial state distribution (P-DDPG-AI).
For P-DDPG-AI we train four models simultaneously and exchange their critics every $K$ iterations.
The $K$ of this parallelization is set to be 20.
Then we choose the best trained model for evaluation.

\begin{figure*}[t!]
    \centering
    \subfigure[InvertedPendulum]{
        \label{fig:ip}
        \includegraphics[width=0.44\linewidth]{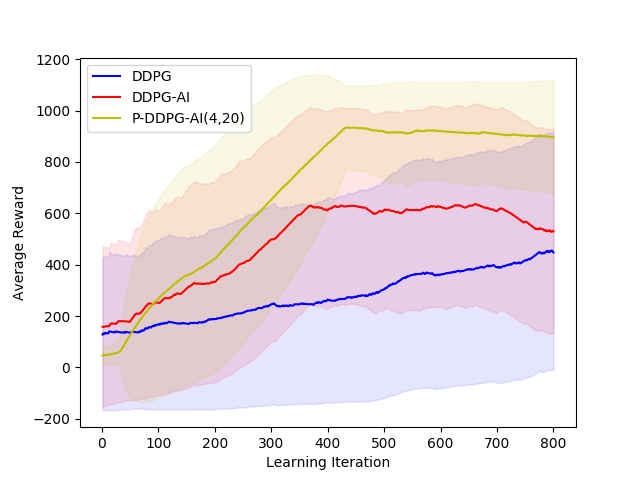}}%
    \qquad
    \subfigure[InvertedDoublePendulum]{
        \label{fig:idp}
        \includegraphics[width=0.44\linewidth]{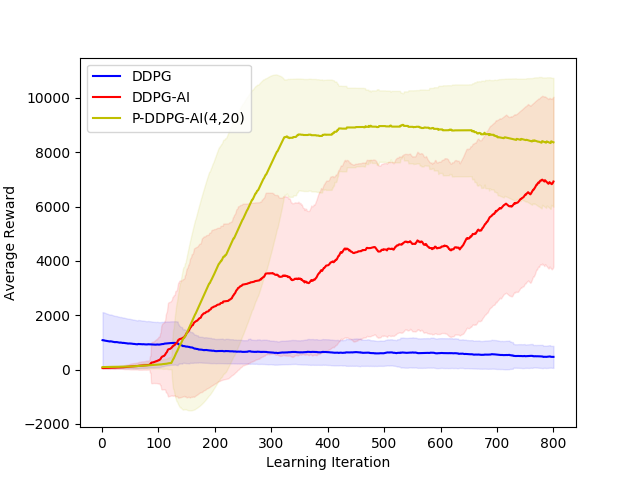}}
    \captionof{figure}{Learning curve for different DDPG variations}
    \vspace{-15pt}
    \label{fig:ddpg}
\end{figure*}

From Fig. \ref{fig:ddpg}, we can observe that DDPG-AI outperforms DDPG due to the adapted initial state distribution, but it is relatively unstable during learning.
However, P-DDPG-AI(4, 20) further improves not only the performance but the stability of training. 
The reason is that once the good learning signals are received, they will be shared and further improve the total performance. 
The results show that the parallelized approach in PRCG works well for the algorithms with adapted initial state distribution.
Since RCG satisfies the condition, it is very suitable to apply the parallelized approach to RCG.

\section{Conclusion and Future Work}
\label{sec:conclusion}

In this paper, a parallelized approach is proposed to utilize the idea of solving tight coupling between a generator and discriminator in GAN to improve the learning of RCG.
The experiments show that our method can make the distribution of learned initial positions more uniform and improve the convergence of the algorithm.
Moreover, this parallelized approach can also be applied to other AC based RL algorithms with adapted initial state distribution.
With multiple models sharing good learning signals and environmental information, the learning is further improved.

During our experiments, we observe that sometimes the fixed exchange rate $K$ will make the learning slightly unstable after some training iteration. The possible reasons are that the critic of one model is exchanged while it has not reached a relatively stable learning or the information contained in the new critic is too different from the previous one. 
For future work, we can make $K$ variable and adaptive to current learning information to improve stability and avoid the parameter searching for $K$.
Another possible future direction is that other ideas used to stabilize GAN training can also be studied for improving AC based RL algorithms.

%===============================================================================

% The maximum paper length is 8 pages excluding references and acknowledgements, and 10 pages including references and acknowledgements

%\clearpage
% The acknowledgments are automatically included only in the final version of the paper.
%\acknowledgments{If a paper is accepted, the final camera-ready version will (and probably should) include acknowledgments. All acknowledgments go at the end of the paper, including thanks to reviewers who gave useful comments, to colleagues who contributed to the ideas, and to funding agencies and corporate sponsors that provided financial support.}

%===============================================================================

\bibliography{example}  % .bib

\end{document}